\documentclass[runningheads]{llncs}

\usepackage{eccv}

\usepackage{xcolor}
\usepackage{colortbl}
\usepackage{eccvabbrv}
\usepackage{units}
\usepackage{graphicx}
\usepackage{booktabs}
\usepackage[section]{placeins}
\usepackage[accsupp]{axessibility}  
\usepackage{subcaption} 
\usepackage{tabularx} 

\usepackage[pagebackref,breaklinks,colorlinks,citecolor=eccvblue]{hyperref}

\usepackage{orcidlink}

\definecolor{yellow}{rgb}{1,1, 0.6}
\definecolor{lightyellow}{rgb}{1,1, 0.8}
\definecolor{orange}{rgb}{1, 0.8, 0.6}
\definecolor{red}{rgb}{1, 0.6, 0.6}
\definecolor{wincolor}{rgb}{0.85, 0.0, 0.0}

\definecolor{darkyellow}{rgb}{0.8, 0.8, 0.5}
\definecolor{darkred}{rgb}{0.7, 0.3, 0.3}
\definecolor{darkgreen}{rgb}{0.3, 0.7, 0.3}
\definecolor{blue}{rgb}{0, 0, 1.0}
\definecolor{green}{rgb}{0, 1.0, 0}
\definecolor{pink}{rgb}{1, 0.4, 0.7}
\begin{document}

\title{SA-GS: Scale-Adaptive Gaussian Splatting for Training-Free Anti-Aliasing}

\titlerunning{SA-GS}

\author{Xiaowei Song*$^{1,2}$ \and
Jv Zheng*$^{1,3}$ \and
Shiran Yuan\inst{1,4}
\and Huan-ang Gao\inst{1} \and Jingwei Zhao\inst{5} \and Xiang He\inst{5} \and Weihao Gu\inst{5} \and Hao Zhao$^{\dagger}$\inst{1}}

\authorrunning{Song et al.}

\institute{Institute for AI Industry Research (AIR), Tsinghua University \and Tongji University \and Ocean University of China \and Duke Kunshan University \and Haomo.ai \\\email{kevin729@tongji.edu.cn}, \email{zhengju@stu.ouc.edu.cn}\\
\email{zhaohao@air.tsinghua.edu.cn}}
\maketitle
\begingroup
\renewcommand\thefootnote{}\footnote{\text{*} Indicates Equal Contribution. \text{\textdagger} Indicates Corresponding Author.
}
\addtocounter{footnote}{1}
\endgroup

\begin{center}{
    \vspace{-1.5em}
    \captionsetup{type=figure}
    \includegraphics[width=1\columnwidth]{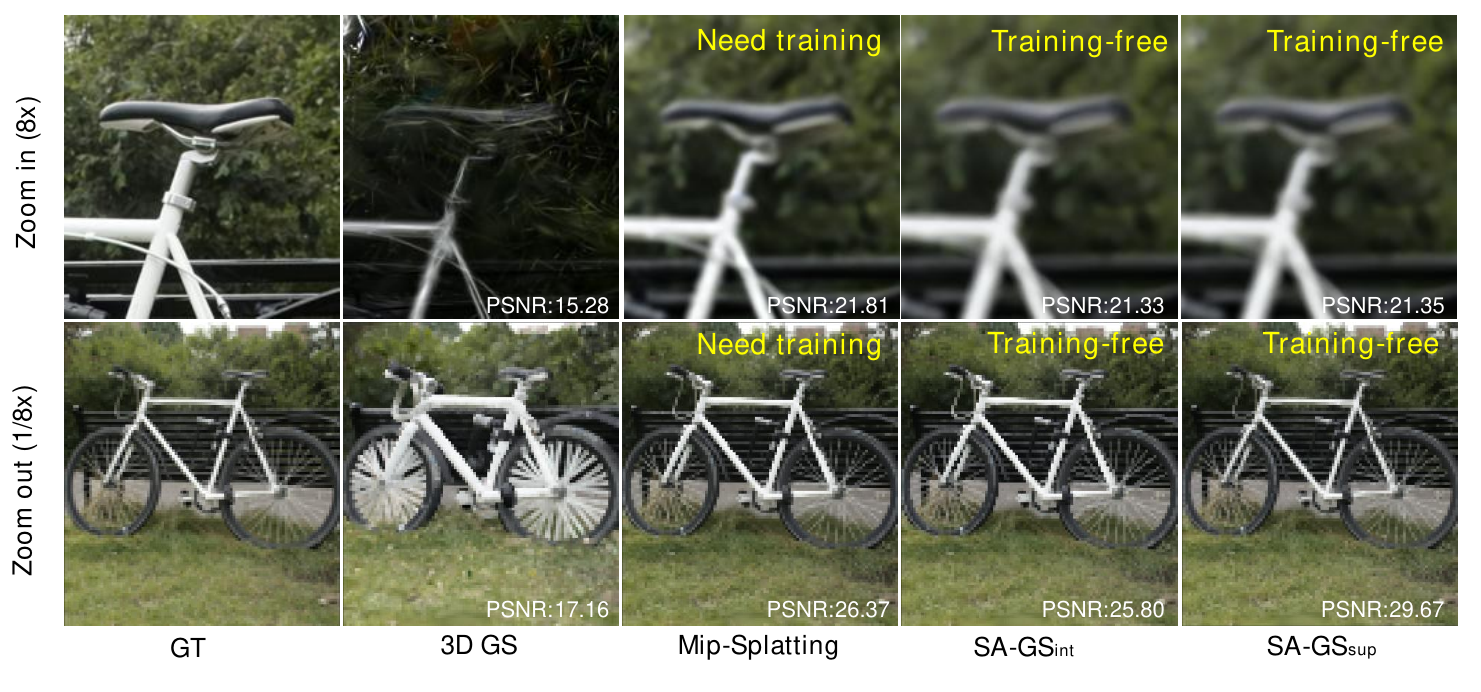}
    \vspace{-2em}
    \captionof{figure}{Under zoom-in, 3D Gaussian Splatting \cite{kerbl20233dgs} ($3DGS$) exhibits significant erosion artefacts, while under zoom-out, it undergoes dramatic dilation. Mip-Splatting \cite{yu2023mip} utilizes 3D smoothing and 2D Mip filters to regularize primitives \textbf{during training}. In contrast, our method is \textbf{training-free} and maintains scale consistency using solely a single \textbf{2D scale-adaptive filter}. Scale adaptation allows us to use \textbf{super-sampling} (named as $SA\text{-}GS_{sup}$ later in the paper) and its limiting case \textbf{integration} (named as $SA\text{-}GS_{int}$ later in the paper) to obtain more accurate results when zooming out.}
    \label{fig:teaser}
}
\end{center}

\begin{abstract}
In this paper, we present a \textbf{S}cale-adaptive method for \textbf{A}nti-aliasing \textbf{G}aussian \textbf{S}platting (SA-GS). While the state-of-the-art method Mip-Splatting needs modifying the training procedure of Gaussian splatting, our method functions at test-time and is training-free. Specifically, SA-GS can be applied to any pretrained Gaussian splatting field as a plugin to significantly improve the field's anti-alising performance. The core technique is to apply 2D scale-adaptive filters to each Gaussian during test time. As pointed out by Mip-Splatting, observing Gaussians at different frequencies leads to mismatches between the Gaussian scales during training and testing. Mip-Splatting resolves this issue using 3D smoothing and 2D Mip filters, which are unfortunately not aware of testing frequency. In this work, we show that a 2D scale-adaptive filter that is informed of testing frequency can effectively match the Gaussian scale, thus making the Gaussian primitive distribution remain consistent across different testing frequencies. When scale inconsistency is eliminated, sampling rates smaller than the scene frequency result in conventional jaggedness, and we propose to integrate the projected 2D Gaussian within each pixel during testing. This integration is actually a limiting case of super-sampling, which significantly improves anti-aliasing performance over vanilla Gaussian Splatting. Through extensive experiments using various settings and both bounded and unbounded scenes, we show SA-GS performs comparably with or better than Mip-Splatting. Note that super-sampling and integration are only effective when our scale-adaptive filtering is activated. Our codes, data and models are available at \url{https://github.com/zsy1987/SA-GS}.
  
\end{abstract}

\section{Introduction}
\label{sec:intro}

Novel View Synthesis (NVS) has played an important role in fields such as visualization\cite{mildenhall2019local}, simulation\cite{wu2023mars,wei2024editable}, automation\cite{zhu2023latitude,zhou2024pad}, and VR/AR\cite{replica19arxiv,peng2023synctalk}. The advent of Neural Radiance Fields (NeRFs)~\cite{mildenhall2020nerf,yuan2023slimmerf} has significantly enhanced the quality of view synthesis while bypassing the need of reconstructing geometry, texture, material and lighting (which is typically a very under-determined inverse problem). Recently, another method, 3D Gaussian Splatting (3DGS)\cite{kerbl20233dgs} has garnered attention from both academia and industry for its high synthesis quality and fast rendering speed. Gaussian primitive-based representation and its corresponding efficient CUDA implementation enable 3DGS\cite{kerbl20233dgs} to render scenes in real-time with intricate details, greatly accelerating NVS systems intended for tasks such as gaming, simulation and multimedia.

\textbf{Problem.} Unlike implicit representations like NeRFs, 3DGS\cite{kerbl20233dgs} utilizes Gaussian primitives to represent 3D scenes in an explicit manner. This is achieved by optimizing the position, scale, transparency, rotation and spherical harmonic coefficients of each Gaussian primitive to fit input images, producing a continuous 3D signal with a complex Gaussian mixture distribution. However, as pointed out by a recent study Mip-Splatting \cite{yu2023mip}, there is a trick in 3DGS not mentioned in the paper, which is introduced to ensure numerical stability. Specifically, 2D dilation is added during training to expand the distribution over a planar region to eliminate the case in which the region is smaller than one single pixel (thus causing instability). This operation guarantees steady updating of the Gaussian primitives, but results in inconsistencies in the degree of dilation and the degree of change in Gaussian scale if the intrinsic and extrinsic parameters of the camera are not equal to the training situation, with artefacts illustrated in Fig. \ref{fig:teaser}. This is due to the fact that 2D dilation is fixed on the pixel space and is not informed of the scale variation of the Gaussian, as illustrated in Fig. \ref{fig:Scale ambiguity}. 

\textbf{Cause \& Solution.} Training data is typically produced using consistent camera settings. Therefore, the use of a fixed dilation operation during the training phase does not result in variations in the dilated scale at the same location. In this case, the Gaussian primitives can still learn a reasonable 2D projection distribution. However, during the rendering phase, the Gaussian scene may be observed at various resolutions and distances. This can compromise the otherwise good 2D projection distribution, resulting in a different $\alpha$-blending process than during the training phase, which ultimately affects the rendering quality. In this paper, we name this phenomenon as \textbf{Gaussian scale mismatch}, which is a property specific to 3DGS and absent in NeRFs. We believe that the 2D projection distribution of Gaussian primitives in the rendering phase should be consistent with the training phase. We correct the 2D dilation operation (in 3DGS) and the 3D smoothing+2D Mip filter (in Mip-Splatting) via a 2D scale-adaptive filter to enforce scale consistency at different rendering parameter settings.

\textbf{Anti-aliasing.} When the projected 2D Gaussian distribution remains consistent with the training at different rendering settings, anti-aliasing is simplified to ensure the synergy of the sampling frequency and the scene frequency. As the sampling frequency decreases, the Nyquist sampling theorem\cite{sainz2004point,nyquist1928certain} may not be satisfied at a certain frequency level, resulting in aliasing effects in the image. Therefore, we introduce conventional anti-aliasing ideas, super-sampling and its limiting case integration, into 3DGS so that the Nyquist sampling theorem is satisfied when zooming out. Notably, super-sampling and integration only make sense after the Gaussian scale mismatch issue is addressed.

\textbf{Significance.} As shown in Fig.~\ref{fig:teaser}, our method can well address the artefacts of vanilla 3DGS while being training-free. Under the zoom-in and zoom-out settings, vanilla 3DGS shows severe visual quality degradation because of erosion and dilation. The quality degradation is actually caused by an intertwined effect of Gaussian scale mismatch and aliasing, which is different from the case that, in NeRFs, artefacts under zoom-in and zoom-out are solely caused by aliasing. While Mip-Splatting \cite{yu2023mip} can address these artefacts, our method is: (1) \textbf{More flexible}. Mip-Splatting needs to modify the training procedure of 3DGS, but our method is a training-free plugin; (2) \textbf{More elegant}. Mip-Splatting resolves the Gaussian scale mismatch issue with 3D smoothing and 2D Mip filter, but our method exploits a single 2D scale-adaptive filter; (3) \textbf{More accurate}. Our method performs comparably with Mip-Splatting but out-performs it under zoom-out because our scale-adaptive formulation can unleash the power of simple and effective strategies super-sampling (and its limiting case integration).

In summary, our contributions are as follows:

\begin{itemize}
    \item[1.] We introduce a training-free approach that can be directly applied to the inference process of any pretrained 3DGS\cite{kerbl20233dgs}  model to resolve its visual artefacts at drastically changed rendering settings. The method itself is named as scale-adaptive Gaussian splatting (SA-GS) as a whole.
    \item[2.] Technically, we propose a 2D scale-adaptive filter that keeps the Gaussian projection scale consistent with the training phase scale at different rendering settings. This scale-adaptive filter also allows simple anti-aliasing techniques (super-sampling and its limiting case integration) to work effectively. 
    \item[3.] Extensive qualitative and quantitative experiments were conducted on the Mip-NeRF 360\cite{barron2022mip} and Blender\cite{mildenhall2020nerf} datasets. Our method achieves superior or comparable performance compared to the state-of-the-art Gaussian anti-aliasing methods, while being training-free.

\end{itemize}

\section{Related Works}

\subsection{Anti-aliased Neural Radiance Fields}

Aliasing is a common issue in conventional computer graphics that occurs when the rendering frequency drastically changes, resulting in visual artefacts such as jagged edges or moiré patterns. These artefacts are caused by the discrete sampling of continuous physical signals. Neural rendering, which is represented by neural radiance fields (NeRFs) \cite{mildenhall2020nerf} or other techniques \cite{chen2019learning, kato2018neural, liu2019soft}, is also troubled by aliasing. Anti-aliasing neural radiance fields are an active research direction, with notable milestones like Mip-NeRF \cite{barron2021mip}, Mip-NeRF 360 \cite{barron2022mip}, Zip-NeRF \cite{barron2023zip} and Tri-MipRF \cite{hu2023tri}. Mip-NeRF \cite{barron2021mip} featurizes the 3d frustum with an approximate ellipse such as the view-dependent distance-aware visual effects are captured by a closed-form geometric encoding of the (approximate) ellipse. Mip-NeRF 360 \cite{barron2022mip} accelerates Mip-NeRF \cite{barron2021mip} with a density proposal network and addresses unbounded scenes using a heuristic contraction rule. Zip-NeRF \cite{barron2023zip} designs a spiral sampling pattern, instead of volume encoding, for the anti-aliasing of radiance fields based on hash grids. Tri-MipRF \cite{hu2023tri} interpolates down-sampled tri-planes (at corresponding scales) to build a mipmap-like representation. Despite various techniques designed for different NeRF backbones (MLP-based, triplane-based or grid-based), most of them are motivated by two old ideas: super-sampling and mipmap. Our method SA-GS is also motivated by super-sampling, but as will be discussed in the next section, in Gaussian splatting fields, other issues need to be addressed before the power of super-sampling can be fully unleashed.

\subsection{Gaussian Splatting}

3D Gaussian Splatting (3DGS) \cite{kerbl20233dgs} is a recently proposed neural\footnote[1]{This is somewhat inaccurate because many people argue that 3DGS does not involve typical neural networks but only matrix multiplication.} rendering paradigm that is primitive-based (like Point-NeRF\cite{Xu2022PointNeRFPN} or ADOP\cite{ruckert2022adop}), uses spherical harmonics as the color representation (like Plenoxels\cite{fridovich2022plenoxels}), and renders at a very fast rate (faster than Instant-NGP\cite{muller2022instant}). While this new paradigm has recently seen much progress in terms of physical integration (PhysGaussian\cite{xie2023physgaussian}) and geometrical alignment (SuGaR\cite{guedon2023sugar}), anti-aliasing for 3DGS\cite{kerbl20233dgs} is not yet well addressed. We note that one fundamental challenge is that when 3DGS\cite{kerbl20233dgs} is rendered at different distances (i.e., resolutions), the Gaussian scale might not match the training-time scale. This mismatch issue entangles with aliasing, making the problem extremely complicated. This has been pointed out by a recent study Mip-Splatting \cite{yu2023mip}, showing that the dilation and erosion issues were caused by this mismatch. In this paper, we demonstrate that classical super-sampling (and its limiting case of integration) is effective for 3DGS\cite{kerbl20233dgs} anti-aliasing, but only under the case of matched Gaussian scales. Unlike the heuristic and problematic Gaussian scale filtering techniques used by 3DGS\cite{kerbl20233dgs} and Mip-Splatting\cite{yu2023mip}, our adaptive solution well addresses the mismatch issue, thus fully unleashing the power of super-sampling for anti-aliasing.

\section{Method}
\begin{figure}[h]
  \centering
  \includegraphics[height=6cm]{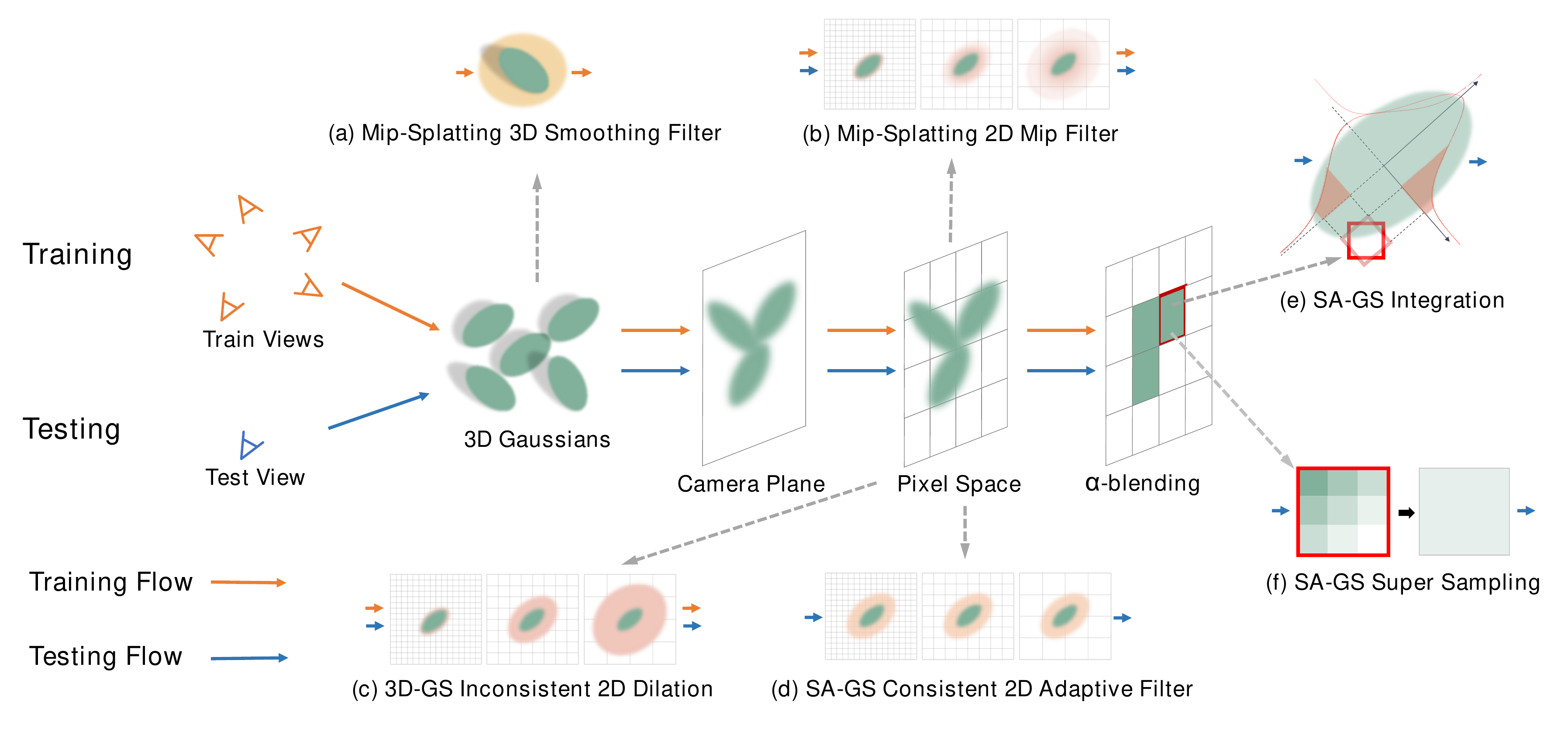}
  \caption{ \textbf{Paradigm Comparison of Gaussian Rasterization Process.} All Gaussian Splatting methods share this framework for training and rendering, but different models use different strategies to process Gaussian primitives. During training, 3DGS\cite{kerbl20233dgs} uses \textbf{(\underline{c})} in pixel space for training stability, but results in scale inconsistencies at different rendering settings; Mip-Splatting utilises \textbf{(\underline{a})} to restrict the Gaussian frequency upper bound in 3D space, and \textbf{(\underline{b})} to emulate box filtering in pixel space. But Mip-Splatting\cite{yu2023mip} still suffers from scale inconsistency and needs to modify the training procedure of 3DGS. Our approach is training-free and only operates on the testing flow. We use \textbf{(\underline{d})} in pixel space to maintain the scale consistency of the Gaussian primitives, and further enhance the anti-aliasing capability of 3DGS by applying \textbf{(\underline{e})} and \textbf{(\underline{f})} to the $\alpha$-blending process. Note that \textbf{(\underline{e})} and \textbf{(\underline{f})} only make sense with \textbf{(\underline{d})} activated.
  }
  \label{fig:gaussian rasterisation process}
\end{figure}

\textbf{A Paradigm comparison} is presented in Fig.~\ref{fig:gaussian rasterisation process}. Overall, our SA-GS method aims to mitigate the artefacts of 3DGS when rendered at different settings (e.g., $8\times$ zoom-in and $1/8\times$ zoom-out shown in Fig.~\ref{fig:teaser}). A notable fact is that in NeRFs \cite{mildenhall2020nerf,barron2021mip,barron2022mip,barron2023zip,hu2023trimiprf}, artefacts under zoom-in/out are caused by aliasing, while for 3DGS these artefacts are caused by the intertwined effects of Gaussian scale mismatch and aliasing. Thus, anti-aliasing techniques make sense only after the Gaussian scale mismatch issue is addressed. The root of Gaussian scale mismatch in vanilla 3DGS, as pointed out by \cite{yu2023mip}, is not described in the paper of \cite{kerbl20233dgs}. The root is shown in Fig.~\ref{fig:gaussian rasterisation process}-(c), which is designed to expand projected 2D Gaussians so that the case of the region is smaller than one single pixel is eliminated. To alleviate the mismatch, Mip-splatting introduces 3D smoothing (Fig.~\ref{fig:gaussian rasterisation process}-(a)) and 2D Mip filter (Fig.~\ref{fig:gaussian rasterisation process}-(b)), during training. Unfortunately, these heuristic methods (Fig.~\ref{fig:gaussian rasterisation process}-(a/b/c)) do not address the mismatch issue in principle, so that conventional anti-alising techniques fail to work for 3DGS and Mip-splatting. Our SA-GS features a 2D scale adaptive filter (Fig.~\ref{fig:gaussian rasterisation process}-(d)) that resolves the mismatch issue in principle and is a training-free plugin. It also unleashes the power of simple anti-aliasing techniques like super-sampling (Fig.~\ref{fig:gaussian rasterisation process}-(f)) and its limiting case integration (Fig.~\ref{fig:gaussian rasterisation process}-(e)) to work for 3DGS. 


\begin{figure}[h]
  \centering
  \includegraphics[height=3.5cm]{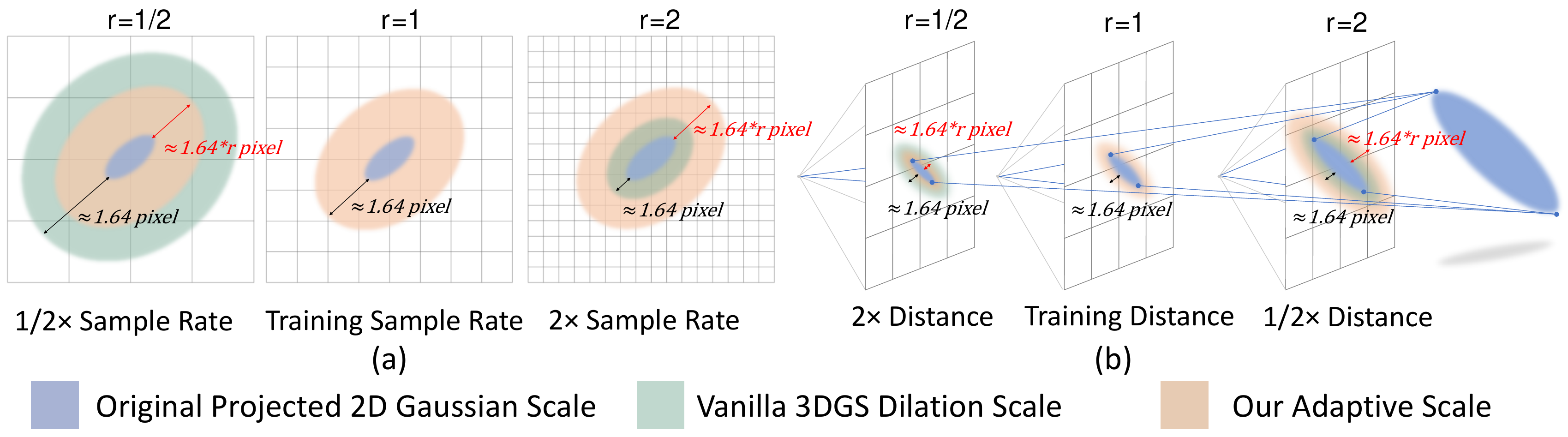}
  \caption{ \textbf{Scale ambiguity.} The heuristic 2D dilation process in vanilla 3DGS code (pointed out by \cite{yu2023mip}) operates on the pixel space and enlarges the projected 2D Gaussian by a fixed amount (around 1.64 pixel). However, a fixed 2D dilation (1.64 pixel) can result in scale ambiguities when representing the same scene at different rendering settings, as shown by the green expansion area. \textbf{(a)} When the Gaussian scale is held constant and the resolution changes, the dilation scale (green) changes inconsistently. \textbf{(b)} When the Gaussian scale changes and the resolution remains constant, the dilation scale (green) does not change with the Gaussian. Our 2D scale-adaptive filter ensures that the Gaussian scale remains consistent across different rendering settings, as shown by the red expansion area. This keeps the scale consistent with the training setup.
  }
  \label{fig:Scale ambiguity}
\end{figure}

\subsection{2D Scale-adaptive Filter}
\label{section:2D Scale Adaptive Filter}
The dilation operation (heuristic and fixed at 1.64 pixel) used by 3DGS\cite{kerbl20233dgs} during training introduces scale ambiguity to the 3D scene, as shown in Fig.~\ref{fig:Scale ambiguity}. As mentioned above, it is crucial to maintain the scale of the Gaussian in the training setup consistent at different rendering settings. 

\begin{figure}[t]
  \centering
  \includegraphics[height=3.7cm]{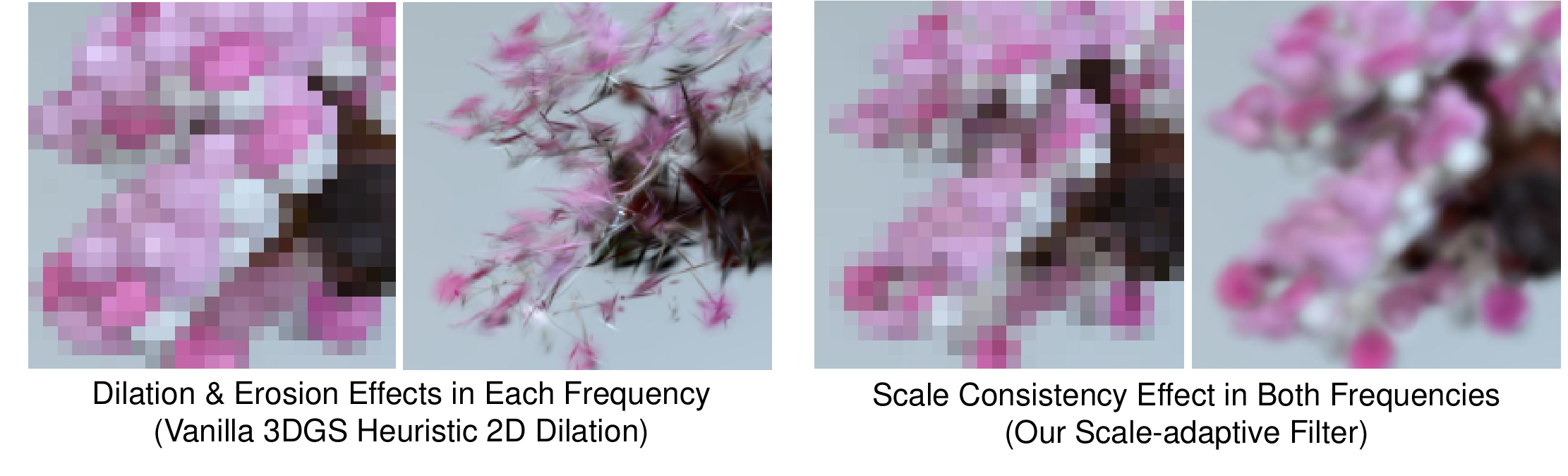}
   \vspace{-1.5em}
  \caption{ \textbf{3DGS heuristic dilation and our scale-adaptive filter.} Our 2D scale-adaptive filter maintains the structure of the scene at any resolution, whereas 3DGS\cite{kerbl20233dgs} dilation (fixed at 1.64 pixel) leads to erroneous dilation at low frequencies and erosion at high frequencies. (Note \emph{dilation} refers to method and artefacts in different contexts.)
  }
  \label{fig:Adaptive filter}
\end{figure}

We propose a 2D scale-adaptive filter that bridges the scale gap between the rendering stage and the training setup, whose effects are shown in Fig. \ref{fig:Adaptive filter}. In pixel space, a 2D Gaussian primitive can be expressed parametrically by its mean $\mathbf{p}_k$ and covariance $\boldsymbol{\Sigma}_k$ as follows:

\begin{equation}
    \begin{aligned}
       \mathcal{G}^{2D}_k(\mathbf{x})=e^{-\frac{1}{2}(\mathbf{x}-\mathbf{p}_k)^T\boldsymbol{\Sigma}_k^{-1}(\mathbf{x}-\mathbf{p}_k)}
    \end{aligned}
\end{equation}

\textbf{Problem.} During the training of vanilla 3DGS, a low-pass Gaussian kernel function $\mathcal{G}_l$ is applied for dilation. This is formally expressed as a convolution between two Gaussians and can eventually be written as $\mathcal{M}_k$:

\begin{equation}
    \begin{aligned}
        \mathcal{G}_k^{2D}(x)_{3DGS} &= \sqrt{\frac{|\boldsymbol{\Sigma}_k+\sigma_l\cdot\mathbf{I}|}{|\boldsymbol{\Sigma}_k|}}(\mathcal{G}_k(\mathbf{p}_k,\boldsymbol{\Sigma}_k) \ast \mathcal{G}_{l}(\mathbf{p}_k,\sigma_l\cdot\mathbf{I}))(x)\\
        &= \mathcal{M}_k(\mathbf{p}_k,\boldsymbol{\Sigma}_k+\sigma_l\cdot\mathbf{I})(x)
    \end{aligned}
\end{equation}

Here, $\sigma_l$ is a fixed hyperparamter that controls the scale of $\mathcal{G}_l$, while $\mathbf{I}$ is 2D unit matrix. Here the problem of 3DGS is that $\sigma_l$ is fixed as 0.3, which approximately leads to $\sqrt{0.3} \times 3 \approx 1.64$ dilation as shown in Fig.~\ref{fig:Scale ambiguity}. 

\textbf{Solution.} During rendering, the scale of the Gaussian primitive in camera space should remain constant, regardless of any changes in rendering frequency. This is achieved by calculating the ratio $r = \frac{\Delta R_p}{\Delta D_c}$. $\Delta R_p$ is the resolution ratio between training and rendering, solving the problem described in Fig. \ref{fig:Scale ambiguity}-(a). $\Delta D_c$ is the distance(focal length) ratio between the rendering camera and the closest orientated training camera, solving the problem described in Fig. \ref{fig:Scale ambiguity}-(b).

\begin{equation}
    \begin{aligned}
       \mathcal{G}_k^{2D}(x,r)_{SA\text{-}GS} &= \sqrt{\frac{|\boldsymbol{\Sigma}_k+\sigma_lr^2\cdot\mathbf{I}|}{|\boldsymbol{\Sigma}_k|}}(\mathcal{G}_k(\mathbf{p}_k,\boldsymbol{\Sigma}_k) \ast r\mathcal{G}_{l}(\mathbf{p}_k,\sigma_l\cdot\mathbf{I}))(x)\\
  &= \mathcal{M}_k(\mathbf{p}_k,\boldsymbol{\Sigma}_k+\sigma_lr^2\cdot\mathbf{I})(x)
    \end{aligned}
\end{equation}

Via this operation (named as 2D scale-adaptive filter), we can ensure a consistent scale and distribution of 2D Gaussian projections in camera space at different rendering settings, such that we can match the training settings.

\subsection{Making Conventional Anti-Aliasing Great Again for Gaussians}
\label{section:Graphics Anti-aliasing applied on Gaussians}

\begin{figure}[t]
  \centering
  \includegraphics[height=5cm]{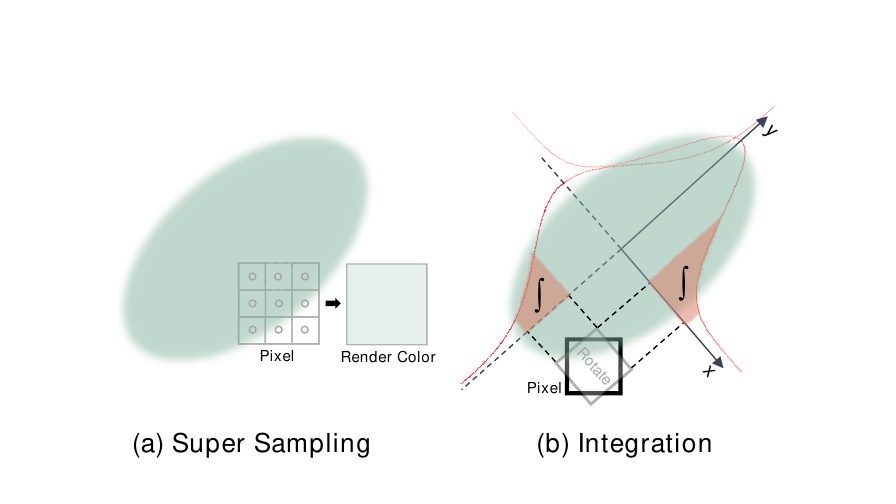}
  \vspace{-2em}
  \caption{ \textbf{Super Sampling and Integration applied on a Gaussian primitive}. Our super sampling method, denoted as (a), involves dividing each pixel into 9 sub-pixels when traversing the order-sorted Gaussians within a tile. Each sub-pixel independently undergoes $\alpha$-blending and weights the Gaussian spherical harmonic coefficient according to the sub-pixel sampling locations. (b) is our integration method that diagonalizes the Gaussian covariance matrix by pixel rotation. This decomposes the integration operation into the product of two marginal Gaussian distributions.
  }
  \label{fig:supersampling_integration}
\end{figure}

Our 2D scale-adaptive filter ensures that the Gaussian distribution remains consistent via matching arbitrary rendering settings with the training setting. Only after this scale adaptation, we can tackle the aliasing issue. Specifically, due to the Nyquist sampling theorem\cite{nyquist1928certain}, the image will show aliasing effects as the rendering frequency decreases. In the conventional graphics literature, there are two techniques that can be used to deal with this problem of aliasing: super-sampling and pre-filtering. 3DGS\cite{kerbl20233dgs} cannot leverage these old techniques to deal with anti-aliasing due to the aforementioned issue of Gaussian scale mismatch. Our method maintains consistent Gaussian scale across different resolutions, allowing for effective removal of scene aliasing. We chose to use super-sampling and its limiting case, integration, instead of pre-filtering, because the pre-filtering affects the $\alpha$-blending procedure of 3DGS which may be a future pursuit.

\subsubsection{\textit{Super-sampling}} Given a pixel $P_t$, when traversing Gaussian primitives that have been order-sorted within a tile, we compute the distance between the centers of the $S\times S$ sub-pixels and the center of the Gaussian primitive separately, as shown in Fig.~\ref{fig:supersampling_integration}(a). These sub-pixels have independent $\alpha$-blending processes and cumulative transparency $T_s$.The color of a pixel $C_t$ is determined by averaging the color of these sampled sub-pixels:

\begin{equation}
    \begin{aligned}
       C_{t} &= \frac{1}{S^2}\sum_{i=1}^G\sum_{s=1}^{S^2}\alpha_s^i\times T_s^i \times \mathcal{F}(SH_i)\\
       T_s^i &= \left\{\begin{array}{lr}
            1, i=1\\
            \prod_{j=1}^{i-1}(1-\alpha_s^j),i>1
        \end{array}
        \right.
    \end{aligned}
\end{equation}
where $G$ is the number of Gaussian primitives on the z-buffer, $\alpha_s^i$ is the opacity calculated based on the distance between $s_{th}$ sub-pixel and $i_{th}$ Gaussian, and $SH_i$ is the spherical harmonic coefficient of the $i_{th}$ Gaussian. The function $\mathcal{F}(\cdot)$ is used for the spherical harmonic coefficient to color conversion. For fast convergence, we set $S = 3$ in all experimental settings.

\begin{figure}[t]
\centering
\includegraphics[width=\textwidth]{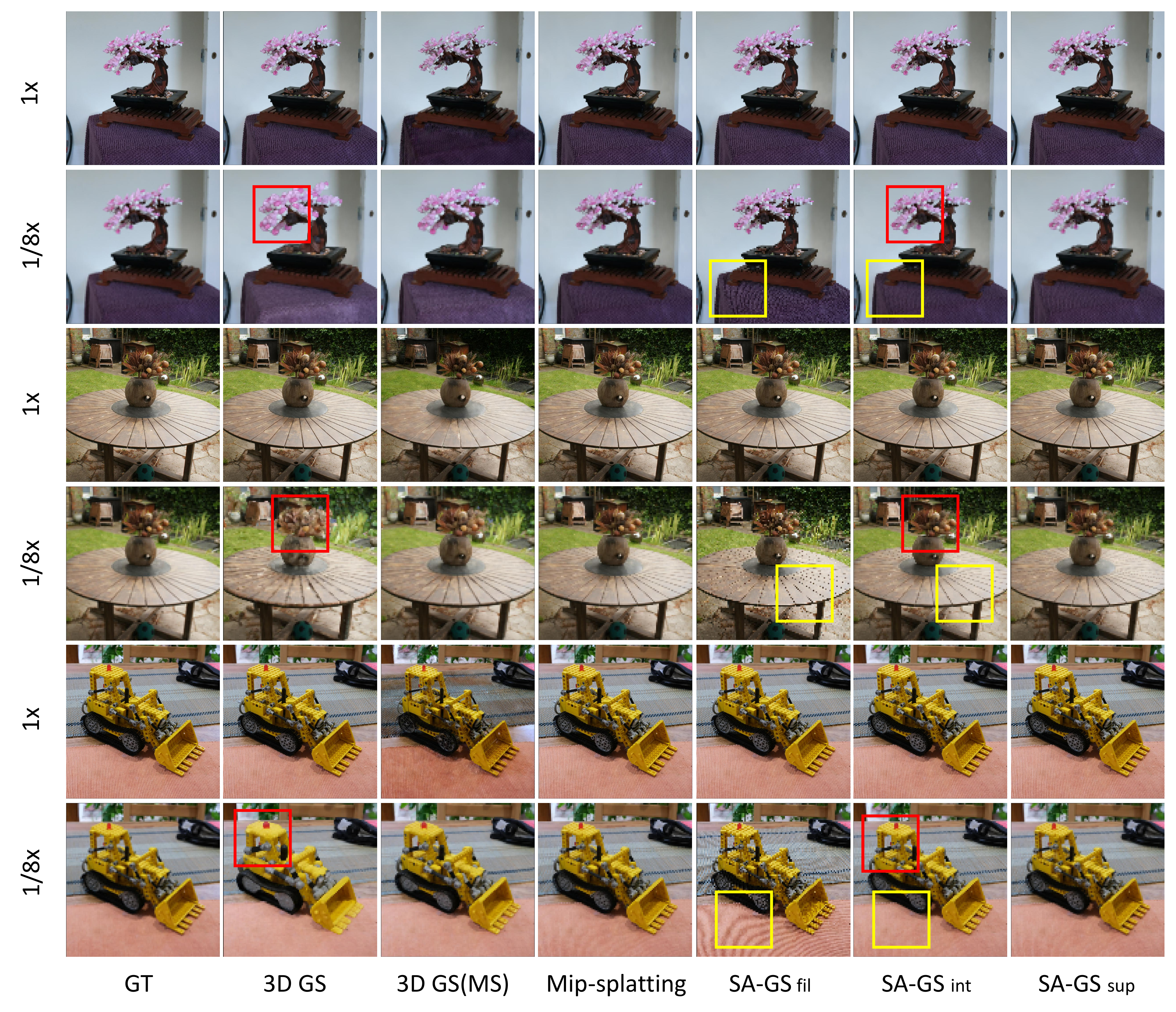}
\vspace{-2em}
\caption{\textbf{Single-scale Training and multi-scale testing on the Mip-NeRF 360 Dataset\cite{barron2022mip} for zoom-out effect.} 3DGS\cite{kerbl20233dgs} has dilation artefacts (red boxes) at low resolutions. Our 2D scale-adaptive filter maintains the Gaussian scale consistency at low resolutions, and the super-sampling and integration methods further remove the aliasing artefacts (yellow boxes), yielding results that surpass Mip-Splatting\cite{yu2023mip}.}
\vspace{-2em}
\label{360-zoomout}
\end{figure}

\subsubsection{\textit{Integration}} When the super-sampling hyperpameter of $S$ goes to infinity, it becomes integration. Consider a single Gaussian's projection on the 2D camera plane, a 2D Gaussian whose PDF we can represent as $f(x,y)$, where $x$ and $y$ are coordinates on the camera plane, and we take axes of the 2D Gaussian Projection as the coordinate system axes, thus making the correlation of the projected Gaussians zero. As the correlation is zero we have that $f(x,y)$ can further be factored into the product $g_x(x)g_y(y)$, where $g(t)=\frac{\exp(-\frac12(\frac{t}{\sigma})^2)}{\sqrt{2\pi}\sigma}$. Let $\Phi(t)=\int_{-\infty}^tg(t)dt$ (subscripts $x$ or $y$ omitted) be the Gaussian integral. Let the region inside the pixel be $P$. Hence when calculating $\alpha$ during the traversal of the Gaussian z-buffer, we need to find the following double integral: 

\begin{equation}
\label{double_int}
\alpha = \iint\limits_Pf(x,y)dxdy
\end{equation}

\textbf{Axis-aligned Case.} When the axes of the pixel are parallel to the 2D Gaussian Projection's axes, the evaluation of (\ref{double_int}) is simple because it can be calculated as the product of two Gaussian single integrals:

\begin{align}[t]
\label{factorize_int}
\iint\limits_Pf(x,y)dxdy&=(\int\limits_{P_x}g_x(x)dx)(\int\limits_{P_y}g_y(y)dy) \nonumber\\
&=(\Phi_x(P_{x\max})-\Phi_x(P_{x\min}))(\Phi_y(P_{y\max})-\Phi_y(P_{y\min}))
\end{align}
where $P_{x\min}$ and $P_{x\max}$ are marginal Gaussian intervals on the x-axis, $P_{y\min}$ and $P_{y\max}$ are the same on the y-axis. However, when the pixel's sides is not aligned with the axes, $x$ and $y$ will be related. In the evaluation of (\ref{double_int}) this is reflected as the inner integral containing variable instead of constant limits, and therefore the factorization carried out in (\ref{factorize_int}) will no longer be applicable.

\textbf{Pixel Rotation.} In our implementation, we solve this problem by \emph{rotating the pixel} such that it aligns with the Gaussian's axes, as shown in Fig.~\ref{fig:supersampling_integration}(b). Thus, the integral can be easily computed using this approach as described in (\ref{factorize_int}). However, rotating the pixel causes a deviation between the integral region and the original pixel region. We prove that for any pixel close enough to the center of the Gaussian to be affected during $\alpha$-blending, there exists a theoretical upper bound for the error. We also verify through numerical experimentation that this is empirically a good approximation. Detailed proofs and experimental results are provided in the supplementary material.

\begin{table*}[h]
    \renewcommand{\tabcolsep}{1pt}
    \centering
    \resizebox{\linewidth}{!}{
        \begin{tabular}{l|ccccp{0.9cm}<{\centering}|ccccp{0.9cm}<{\centering}|ccccp{0.9cm}<{\centering}}
        & \multicolumn{5}{c|}{PSNR $\uparrow$} & \multicolumn{5}{c|}{SSIM $\uparrow$} & \multicolumn{5}{c}{LPIPS$\downarrow$}\\
              & 1 Res. & $\nicefrac{1}{2}$ Res. & $\nicefrac{1}{4}$ Res. & $\nicefrac{1}{8}$ Res. & Avg. & 1 Res. & $\nicefrac{1}{2}$ Res. & $\nicefrac{1}{4}$ Res. & $\nicefrac{1}{8}$ Res. & Avg. & 1 Res. & $\nicefrac{1}{2}$ Res. & $\nicefrac{1}{4}$ Res. & $\nicefrac{1}{8}$ Res & Avg. \\ \hline

3DGS\cite{kerbl20233dgs} 
& \cellcolor{red}{29.26} & 26.84 & 22.16 & 19.63 & 24.47
& \cellcolor{red}{0.877} & 0.863 & 0.726 & 0.612 & 0.769
& \cellcolor{red}{0.185} & 0.148	& 0.198 & 0.223 & 0.189
\\
3DGS(MS)\cite{kerbl20233dgs}
& 20.11 & 23.50 & \cellcolor{red}{32.51} & 23.72 & 24.96
& 0.604 & 0.774 & \cellcolor{red}{0.956} & 0.832 & 0.792
& 0.389 & 0.212 & \cellcolor{red}{0.051} & 0.118 & 0.192
\\
Mip-Splatting\cite{yu2023mip}
& \cellcolor{red}{29.26} & \cellcolor{orange}{30.23} & \cellcolor{yellow}{30.56} & \cellcolor{orange}{29.61} & \cellcolor{orange}{29.91}
& 0.875 & \cellcolor{orange}{0.909} & \cellcolor{yellow}{0.929} & \cellcolor{orange}{0.934} & \cellcolor{orange}{0.911}
& 0.187 & \cellcolor{orange}{0.116} & \cellcolor{yellow}{0.080} & \cellcolor{orange}{0.066} & \cellcolor{orange}{0.113}
\\
\hline
$SA\text{-}GS_{fil}\text{(ours)}$
& \cellcolor{red}{29.26} & 29.80 & 28.29 & 25.58 & 28.23
& \cellcolor{red}{0.877} & 0.901 & 0.875 & 0.809 & 0.866
& \cellcolor{red}{0.185} & 0.123 & 0.126 & 0.171 & 0.151
\\
$SA\text{-}GS_{int}\text{(ours)}$
& 29.14	& \cellcolor{yellow}{30.06} & 30.13	& \cellcolor{yellow}{28.81} & \cellcolor{yellow}{29.53}
& 0.873	& \cellcolor{yellow}{0.905} & 0.921 & \cellcolor{yellow}{0.919} & \cellcolor{yellow}{0.904}
& 0.188 & \cellcolor{yellow}{0.118} & 0.086 & \cellcolor{yellow}{0.078} & \cellcolor{yellow}{0.118}
\\
$SA\text{-}GS_{sup}\text{(ours)}$
& \cellcolor{red}{29.26} & \cellcolor{red}{30.45} & \cellcolor{orange}{31.75} & \cellcolor{red}{32.53} & \cellcolor{red}{31.00}
& \cellcolor{yellow}{0.876} & \cellcolor{red}{0.912} & \cellcolor{orange}{0.938} & \cellcolor{red}{0.951} & \cellcolor{red}{0.919}
& \cellcolor{yellow}{0.186} & \cellcolor{red}{0.114} & \cellcolor{orange}{0.073} & \cellcolor{red}{0.053} & \cellcolor{red}{0.106}
\end{tabular}}
    \vspace{0.5em}
    \caption{
        \textbf{Single-scale training and multi-scale testing on the Mip-NeRF 360 Dataset\cite{barron2022mip}.} Except for $3DGS(MS)$\cite{kerbl20233dgs}, which is trained on multiple scales, all other methods are trained on the largest scale (1×) and evaluated across four scales (1×, $\nicefrac{1}{2}$×, $\nicefrac{1}{4}$×, and $\nicefrac{1}{8}$×), simulating zoom-out effect. $SA\text{-}GS_{fil}$ means we only use the 2D scale-adaptive filter. All our variants significantly surpass $3DGS$ at low resolutions, and $SA\text{-}GS_{sup}$ yields better results than $Mip\text{-}Splatting$\cite{yu2023mip}.
    }
    \vspace{-2em}
    \label{tab:avg_single360_zoom_out_results}
\end{table*}

\section{Experiments}

We first present the implementation details of SA-GS. We then evaluate its performance on the the unbounded Mip-NeRF 360 dataset\cite{barron2022mip} and the bounded Blender dataset\cite{mildenhall2020nerf}. Finally, we discuss some limitations of our approach.

\subsection{Implementation}

In our SA-GS framework, an advantage over Mip-splatting is that modifications are exclusively performed during the testing phase of 3DGS\cite{kerbl20233dgs}. Therefore, in the training phase, we kept all the settings of the original Gaussian model, including training rounds, hyperparameter settings and densification strategy. In our super-sampling method, we synchronize all pixel threads\footnote{In 3DGS, each pixel is associated with a thread.} within the same block before performing the alpha calculation to initialize the cumulative transparency in the shared memory. In our integration method, we project the pixel corner points towards the Gaussian axis to obtain an interval of Gaussian distribution for the marginal distributions. To simulate the pixel rotation, we multiply the pixel region area by a weight of $\frac{1}{sin\theta+cos\theta}$ to compensate the error caused by rotation, where $\theta$ is the angle between the long Gaussian axis and the x-axis of the pixel plane. Please refer to the supplementary material for details.

\subsection{Evaluation on the Mip-NeRF 360 Dataset}

\subsubsection{Single-scale Training and Multi-scale Testing:}We simulated the effect of zoom-out and zoom-in on this dataset and retrained all the baseline models for comparison. The same setup of Mip-Splatting\cite{yu2023mip} is used. Specifically, for zoom-out, we trained 3DGS\cite{kerbl20233dgs} using full resolution images and then plug in our method to test on progressively lower resolutions (1×, $\nicefrac{1}{2}$×, $\nicefrac{1}{4}$×, $\nicefrac{1}{8}$×). For zoom-in, we trained 3DGS\cite{kerbl20233dgs} using $\nicefrac{1}{8}$× resolution images and then plug in our method to test on progressively higher resolutions (1×, 2×, 4×, 8×).  Table \ref{tab:avg_single360_zoom_out_results} and Table \ref{tab:avg_single360_zoom_in_results} show the quantitative results for these two experiments.

When \textbf{zooming out}, our method gives comparable results at the training resolution and better performance at lower resolutions. The (heuristic and fixed) dilation operation of 3DGS\cite{kerbl20233dgs} leads to severe degradation at low resolutions. Mip-Splatting\cite{yu2023mip} replaces 3DGS's dilation operation with 3D smoothing and a 2D Mip filter, but still suffers from scale ambiguity at different resolutions. As depicted in Figure \ref{360-zoomout}, our 2D scale-adaptive filter ensures that the Gaussian distribution is consistent with the training settings. Our integration and super-sampling modules further enhance the anti-aliasing effect. The super-sampling version of our method gives the best results for this setting, which exceeds Mip-Splatting\cite{yu2023mip} by \textbf{1.1dB} in PSNR (see Tab.~\ref{tab:avg_single360_zoom_out_results}).

\begin{figure}[t]
\centering
\includegraphics[width=0.6\textwidth]{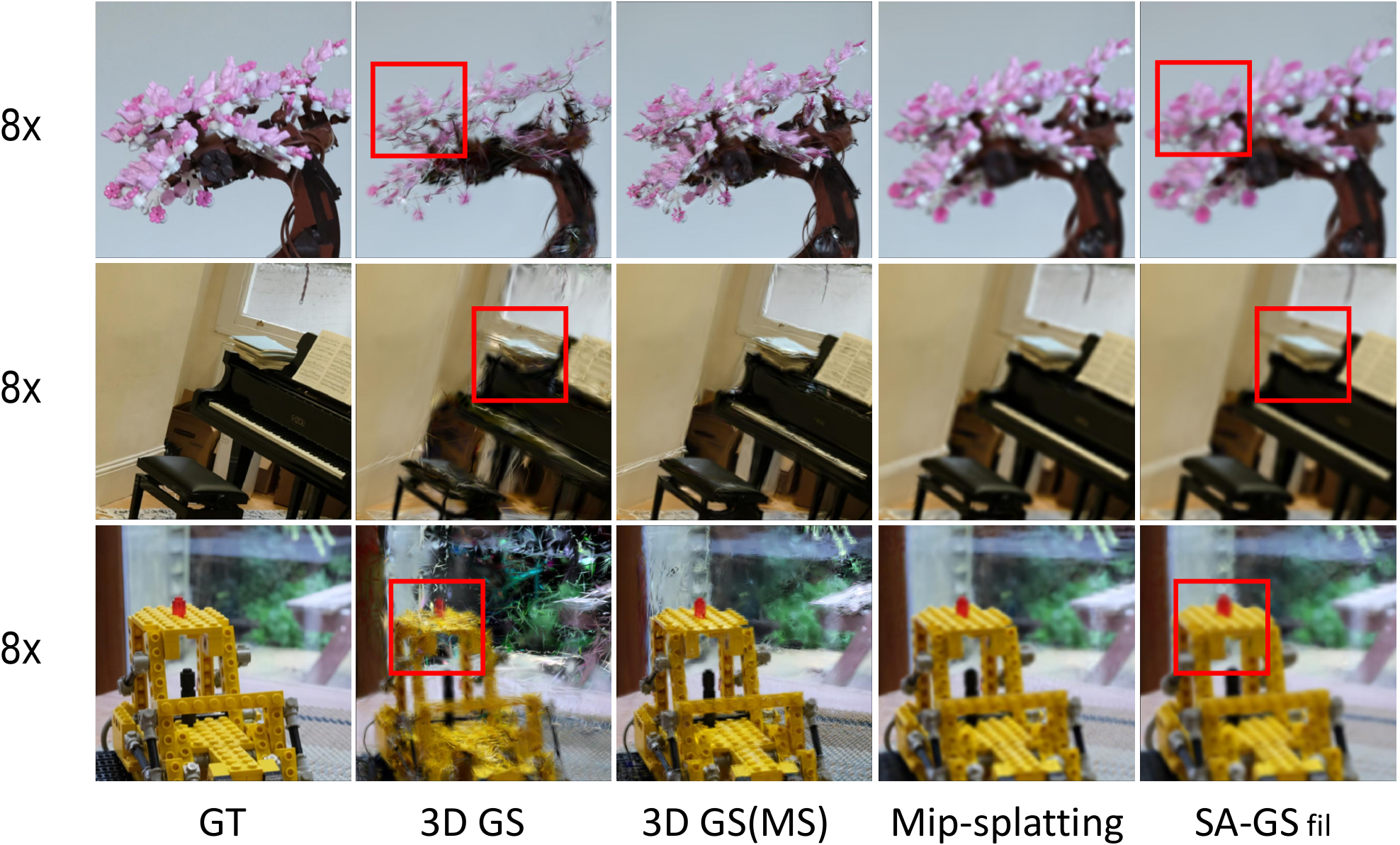}

\caption{\textbf{Single-scale Training and multi-scale testing on the Mip-NeRF 360 Dataset\cite{barron2022mip} for zoom-in effect.} 3DGS\cite{kerbl20233dgs} sees erosion artefacts (red boxes) at high resolution. By using only 2D scale-adaptive filters, we achieve a stable quality improvement at high resolutions without any re-training.}
\label{360-zoomin}
\end{figure}

When \textbf{zooming in}, 3DGS\cite{kerbl20233dgs} shows severe erosion artefacts at high resolution. Mip-Splatting\cite{yu2023mip} uses the 3D smoothing filter to reduce the effects of scale ambiguity of 2D filtering. As depicted in Figure \ref{360-zoomin}, our method keeps the Gaussian scales consistent and greatly reduces erosion artefacts. Note that integration and super sampling are only designed to address the decrease in sampling frequency (zoom-out). The most significant contribution is made by 2D scale-adaptive filter, which produces results comparable to Mip-Splatting\cite{yu2023mip}.
\vspace{-1em}

\vspace{1em}
\begin{table*}[h]
    \renewcommand{\tabcolsep}{1pt}
    \small
    \centering
    \resizebox{\linewidth}{!}{
        \begin{tabular}{l|ccccp{0.9cm}<{\centering}|ccccp{0.9cm}<{\centering}|ccccp{0.9cm}<{\centering}}
        & \multicolumn{5}{c|}{PSNR $\uparrow$} & \multicolumn{5}{c|}{SSIM $\uparrow$} & \multicolumn{5}{c}{LPIPS$\downarrow$}\\
              & 1 Res. & 2 Res. & 4 Res. & 8 Res. & Avg. & 1 Res. & 2 Res. & 4 Res. & 8 Res. & Avg. & 1 Res. & 2 Res. & 4 Res. & 8 Res & Avg. \\ \hline

3DGS\cite{kerbl20233dgs} 
& \cellcolor{orange}{33.96}	& 22.47 & 18.69 & 17.32 & 23.11
& \cellcolor{orange}{0.974}	& 0.747 & 0.514 & 0.460 & 0.674
& \cellcolor{orange}{0.028} & 0.204	& 0.410 & 0.504 & 0.286
\\
3DGS(MS)\cite{kerbl20233dgs}
& 23.72 & \cellcolor{red}{32.51} & \cellcolor{yellow}{23.50} & \cellcolor{yellow}{20.11} & \cellcolor{yellow}{24.96}
& 0.832 & \cellcolor{red}{0.956} & \cellcolor{red}{0.774} & \cellcolor{yellow}{0.604} & \cellcolor{orange}{0.792}
& 0.118 & \cellcolor{red}{0.051} & \cellcolor{red}{0.212} & \cellcolor{red}{0.389} & \cellcolor{red}{0.192}
\\
Mip-Splatting\cite{yu2023mip}
& \cellcolor{red}{34.62} & \cellcolor{orange}{28.86} & \cellcolor{red}{25.99} & \cellcolor{red}{24.95} & \cellcolor{red}{28.60}
& \cellcolor{red}{0.977} & \cellcolor{orange}{0.872} & \cellcolor{orange}{0.718} & \cellcolor{red}{0.641}  & \cellcolor{red}{0.802}
& \cellcolor{red}{0.025} & \cellcolor{orange}{0.154} & \cellcolor{orange}0.318 & \cellcolor{orange}0.430 & \cellcolor{orange}0.232
\\

\hline
$SA\text{-}GS_{fil}\text{(ours)}$
& \cellcolor{orange}{33.96}	& \cellcolor{yellow}{27.89}	& \cellcolor{orange}{25.32}	& \cellcolor{orange}{24.40}	& \cellcolor{orange}{27.89}
& \cellcolor{orange}{0.974}	& \cellcolor{yellow}{0.840}	& \cellcolor{yellow}{0.677}	& \cellcolor{orange}{0.615}	& \cellcolor{yellow}{0.777}
& \cellcolor{orange}{0.028}	& \cellcolor{yellow}0.189	& \cellcolor{yellow}0.360	& \cellcolor{yellow}0.465	& \cellcolor{yellow}0.260
\\
\end{tabular}
    }
    \vspace{0.5em}
    \caption{
        \textbf{Single-scale training and multi-scale testing on the Mip-NeRF 360 Dataset\cite{barron2022mip}.} Except for $3DGS(MS)$\cite{kerbl20233dgs}, which is trained on multiple scales, all other re-training methods are trained on the \nicefrac{1}{8} scale (1×) and evaluated across four scales (1×, 2×, 4×, and 8×), simulating zoom-in effect. $SA\text{-}GS_{fil}$ means we only use 2D Scale-adaptive Filter. Our method significantly surpasses $3DGS$\cite{kerbl20233dgs} at high resolutions and produce results comparable to $Mip\text{-}Splatting$\cite{yu2023mip}. Note that the performance of $SA\text{-}GS_{fil}$ is achieved without re-training.
    }
    \vspace{-2em}
    \label{tab:avg_single360_zoom_in_results}
\end{table*}

\vspace{-2em}
\subsection{Evaluation on the Blender Dataset}
\subsubsection{Multi-scale Training and Multi-scale Testing:} Following Mip-splatting, all baseline models were trained using multi-scale data from the $train+test$ section of the dataset and evaluated with multiscale data from the $val$ section. We follow the image sampling ratio in Mip-Splatting\cite{yu2023mip} to train the 3DGS\cite{kerbl20233dgs}. Our quantitative evaluation is shown in Table \ref{tab:avg_multiblender_results}. Our  approach yields comparable results with Mip-Splatting\cite{yu2023mip}, and we use vanilla 3DGS\cite{kerbl20233dgs} on multi-scale training only and do not need to modify the training procedure. Meanwhile, our approach significantly outperforms 3DGS\cite{kerbl20233dgs}, demonstrating stable performance at different resolutions. 3DGS\cite{kerbl20233dgs} performance degrades as resolution decreases, even in the case that it is trained on multi-scale.

\begin{figure}
\centering
\includegraphics[height=8cm]{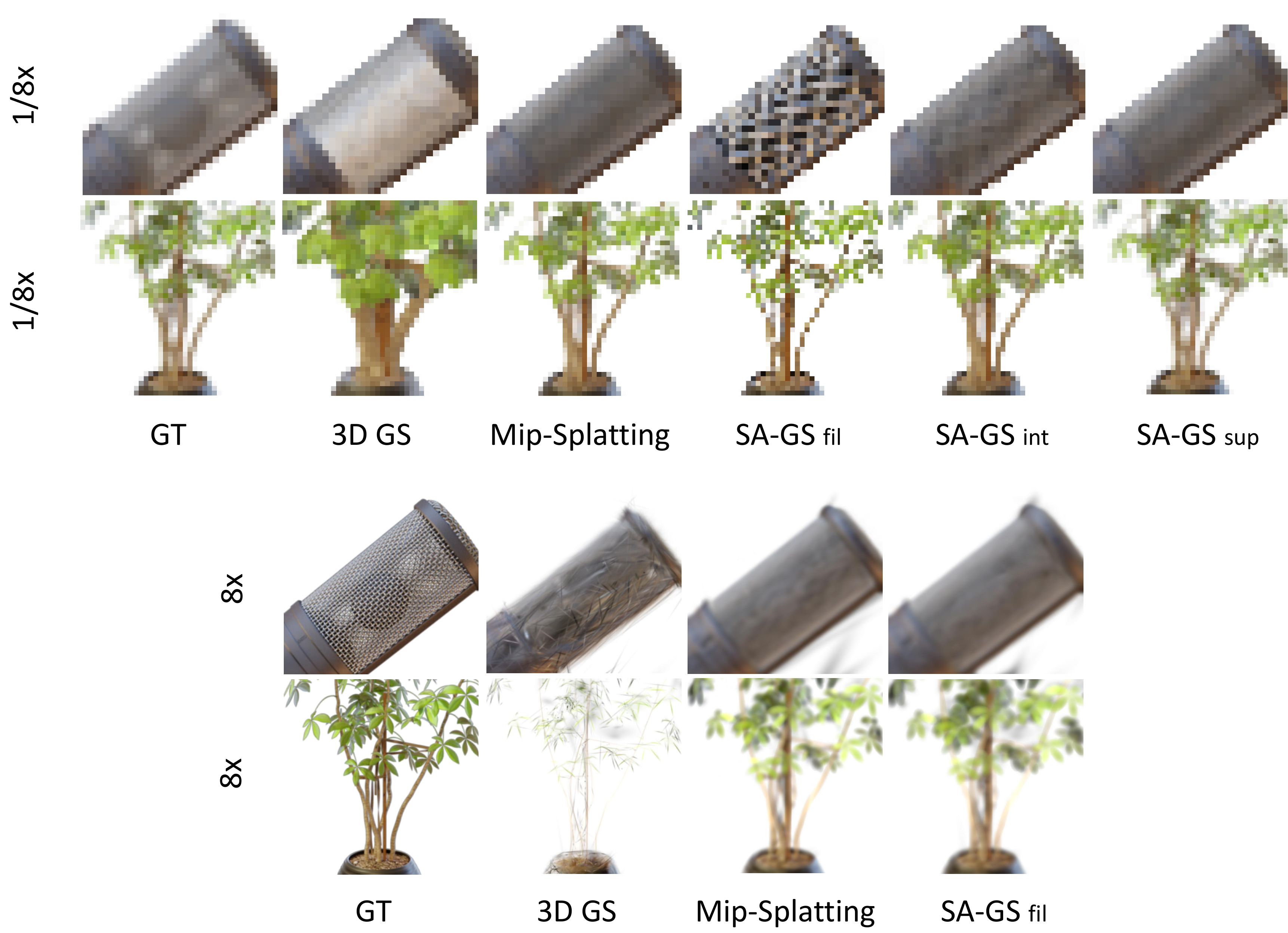}
\vspace{-1em}
\caption{\textbf{Single-scale Training and multi-scale testing on the Blender\cite{mildenhall2020nerf} Dataset for zoom-in and zoom out effect}. Our 2D scale-adaptive filter maintains the consistency of the 2D Gaussian projection when zooming out. Moreover, it alleviates erosion artefacts and does not modify the training procedure when zooming in. We use super-sampling and integration methods to further address aliasing.}
\label{blender}
\end{figure}

\begin{table*}
    \renewcommand{\tabcolsep}{1pt}
    \centering
    \resizebox{\linewidth}{!}{
        \begin{tabular}{l|ccccp{0.9cm}<{\centering}|ccccp{0.9cm}<{\centering}|ccccp{0.9cm}<{\centering}}
        & \multicolumn{5}{c|}{PSNR $\uparrow$} & \multicolumn{5}{c|}{SSIM $\uparrow$} & \multicolumn{5}{c}{LPIPS$\downarrow$}\\
              & 1 Res. & $\nicefrac{1}{2}$ Res. & $\nicefrac{1}{4}$ Res. & $\nicefrac{1}{8}$ Res. & Avg. & 1 Res. & $\nicefrac{1}{2}$ Res. & $\nicefrac{1}{4}$ Res. & $\nicefrac{1}{8}$ Res. & Avg. & 1 Res. & $\nicefrac{1}{2}$ Res. & $\nicefrac{1}{4}$ Res. & $\nicefrac{1}{8}$ Res & Avg. \\ \hline

3DGS\cite{kerbl20233dgs}
& \cellcolor{orange}31.51 & 32.66 & 31.21 & 28.25 & 30.91
& \cellcolor{orange}0.962 & \cellcolor{orange}0.972 & 0.968 & 0.945 & 0.962
& \cellcolor{orange}0.050 & \cellcolor{orange}0.031 & 0.030 & 0.045 & 0.039
\\
Mip-Splatting\cite{yu2023mip}
&\cellcolor{red}32.81 & \cellcolor{red}34.49 & \cellcolor{red}35.45& \cellcolor{orange}35.50 & \cellcolor{red}34.56 
&\cellcolor{red}0.967 & \cellcolor{red}0.977& \cellcolor{red}0.983& \cellcolor{red}0.988 & \cellcolor{red}0.979
&\cellcolor{red}0.035 & \cellcolor{red}0.019& \cellcolor{red}0.013& \cellcolor{red}0.010& \cellcolor{red}0.019
\\

\hline

$SA\text{-}GS_{int}\text{(ours)}$
& \cellcolor{yellow}30.84 & \cellcolor{orange}32.71 & \cellcolor{yellow}34.26 & \cellcolor{yellow}32.80 & \cellcolor{yellow}32.65
& \cellcolor{yellow}0.956 & \cellcolor{yellow}0.969 & \cellcolor{yellow}0.978 & \cellcolor{yellow}0.979 & \cellcolor{yellow}0.971
& \cellcolor{yellow}0.055 & \cellcolor{orange}0.031 & \cellcolor{yellow}0.021 & \cellcolor{yellow}0.019 & \cellcolor{yellow}0.032
\\
$SA\text{-}GS_{sup}\text{(ours)}$
& 30.80 & \cellcolor{yellow}32.67 & \cellcolor{orange}35.06 & \cellcolor{red}35.77 & \cellcolor{orange}33.58
& \cellcolor{yellow}0.956 & \cellcolor{yellow}0.969 & \cellcolor{orange}0.980 & \cellcolor{orange}0.985 & \cellcolor{orange}0.973
& 0.056 & 0.032 & \cellcolor{orange}0.020 & \cellcolor{orange}0.014 & \cellcolor{orange}0.031
\end{tabular}
    }
    \vspace{0.5em}
    \caption{
        \textbf{Multi-scale Training and Multi-scale Testing on the Blender dataset\cite{mildenhall2020nerf}.} Our training-free approach yields comparable results compared to Mip-Splatting\cite{yu2023mip}. Meanwhile, our approach significantly outperforms $3DGS$\cite{kerbl20233dgs} and demonstrates stable performance at different resolutions.
    }
    \label{tab:avg_multiblender_results}
\end{table*}

\begin{table*}
    \renewcommand{\tabcolsep}{1pt}
    \centering
    \resizebox{\linewidth}{!}{
        \begin{tabular}{l|ccccp{0.9cm}<{\centering}|ccccp{0.9cm}<{\centering}|ccccp{0.9cm}<{\centering}}
        & \multicolumn{5}{c|}{PSNR $\uparrow$} & \multicolumn{5}{c|}{SSIM $\uparrow$} & \multicolumn{5}{c}{LPIPS$\downarrow$}\\
              & 1 Res. & $\nicefrac{1}{2}$ Res. & $\nicefrac{1}{4}$ Res. & $\nicefrac{1}{8}$ Res. & Avg. & 1 Res. & $\nicefrac{1}{2}$ Res. & $\nicefrac{1}{4}$ Res. & $\nicefrac{1}{8}$ Res. & Avg. & 1 Res. & $\nicefrac{1}{2}$ Res. & $\nicefrac{1}{4}$ Res. & $\nicefrac{1}{8}$ Res & Avg. \\ \hline

3DGS 
& \cellcolor{red}35.10 & 27.91 & 22.42 & 18.76 & 26.05
& \cellcolor{red}0.974 & 0.949 & 0.862 & 0.736 & 0.880
& \cellcolor{red}0.029 & 0.033 & 0.069 & 0.133 & 0.066
\\
3DGS(MS)
& 31.51 & 32.66 & \cellcolor{yellow}31.21 & \cellcolor{orange}28.25 & 30.91
& 0.962 & 0.972 & 0.968 & 0.945 & 0.962
& 0.050 & 0.031 & 0.030 & 0.045 & 0.039
\\
Mip-Splatting
& \cellcolor{yellow}34.59 & \cellcolor{orange}35.11 & \cellcolor{orange}31.98 & \cellcolor{yellow}28.14 & \cellcolor{orange}32.45
& \cellcolor{orange}0.973 & \cellcolor{orange}0.979 & \cellcolor{orange}0.975 & \cellcolor{orange}0.952 & \cellcolor{orange}0.970
& \cellcolor{yellow}0.032 & \cellcolor{orange}0.019 & \cellcolor{orange}0.019 & \cellcolor{orange}0.029 & \cellcolor{orange}0.025
\\

\hline
$SA\text{-}GS_{fil} \text{(ours)}$
& \cellcolor{orange}34.60 & 34.33 & 31.02 & 27.59 & \cellcolor{yellow}31.89
& \cellcolor{orange}0.973 & 0.977 & 0.968 & \cellcolor{yellow}0.947 & \cellcolor{yellow}0.966
& \cellcolor{orange}0.031 & 0.022 & 0.036 & 0.067 & 0.039
\\
$SA\text{-}GS_{int} \text{(ours)}$
& 34.35 & \cellcolor{yellow}34.39 & 30.99 & 26.89 & 31.65
& 0.972 & \cellcolor{yellow}0.978 & \cellcolor{yellow}0.971 & 0.940 & 0.965
& \cellcolor{yellow}0.032 & \cellcolor{yellow}0.020 & \cellcolor{yellow}0.023 & \cellcolor{yellow}0.039 & \cellcolor{yellow}0.029
\\
$SA\text{-}GS_{sup} \text{(ours)}$
& 34.49 & \cellcolor{red}36.58 & \cellcolor{red}37.50 & \cellcolor{red}35.64 & \cellcolor{red}36.06
& 0.972 & \cellcolor{red}0.980 & \cellcolor{red}0.985 & \cellcolor{red}0.985 & \cellcolor{red}0.981
& \cellcolor{yellow}0.032 & \cellcolor{red}0.018 & \cellcolor{red}0.014 & \cellcolor{red}0.013 & \cellcolor{red}0.019
\end{tabular}
    }
    \vspace{0.5em}
    \caption{
            \textbf{Single-scale training and Multi-scale testing on the Blender dataset\cite{mildenhall2020nerf} for zoom-out effect.} We use the same experiment protocol and model naming with the Mip-NeRF 360\cite{barron2022mip} experiment (of Table~\ref{tab:avg_single360_zoom_out_results}). Our method outperforms 3DGS\cite{kerbl20233dgs}, while $SA\text{-}GS_{sup}$ significantly surpasses all previous works.
    }
    \vspace{-0.5em}
    \label{tab:avg_singleblender_zoom_out_results}
\end{table*}

\subsubsection{Single-scale Training and Multi-scale Testing:}We maintain a experiment protocol consistent with the Mip-NeRF 360\cite{barron2022mip} experiment mentioned above (Section 4.2) to evaluate the zoom-out and zoom-in effects. We also keep the same data split as in the multi-scale training experiment described above. Table \ref{tab:avg_singleblender_zoom_out_results} and Table \ref{tab:avg_singleblender_zoom_in_results} show the quantitative results for zoom-out and zoom-in effects. The qualitative results are shown in Fig. \ref{blender}.

\textbf{For zoom-out}, our method achieves performance close to 3DGS\cite{kerbl20233dgs} at training resolution and a steady increase in performance at lower resolutions. The super-sampling version of our method $SA\text{-}GS_{sup}$ significantly outperforms Mip-Splatting\cite{yu2023mip} to achieve performance with a gain of \textbf{3.61dB} in PSNR in this setting (see Tab.~\ref{tab:avg_singleblender_zoom_out_results}). \textbf{For zoom-in}, Our 2D scale-adaptive filter achieves comparable results to Mip-Splatting\cite{yu2023mip}, and our method is training-free.

\begin{table*}
    \renewcommand{\tabcolsep}{1pt}
    \centering
    \resizebox{\linewidth}{!}{
        \begin{tabular}{l|ccccp{0.9cm}<{\centering}|ccccp{0.9cm}<{\centering}|ccccp{0.9cm}<{\centering}}
        & \multicolumn{5}{c|}{PSNR $\uparrow$} & \multicolumn{5}{c|}{SSIM $\uparrow$} & \multicolumn{5}{c}{LPIPS$\downarrow$}\\
              & 1 Res. & 2 Res. & 4 Res. & 8 Res. & Avg. & 1 Res. & 2 Res. & 4 Res. & 8 Res. & Avg. & 1 Res. & 2 Res. & 4 Res. & 8 Res & Avg. \\ \hline

3DGS\cite{kerbl20233dgs}
& \cellcolor{red}36.97 & 24.33 & 21.01 & 19.63 & 25.44
& \cellcolor{red}0.988 & 0.886 & 0.820 & 0.821 & 0.879
& \cellcolor{red}0.013 & 0.065 & 0.130 & 0.159 & 0.092
\\
3DGS(MS)\cite{kerbl20233dgs}
& 28.25 & \cellcolor{red}31.21 & \cellcolor{red}32.66 & \cellcolor{red}31.51 & \cellcolor{red}30.91
& 0.945 & \cellcolor{red}0.968 & \cellcolor{red}0.972 & \cellcolor{red}0.962 & \cellcolor{red}0.962
& 0.045 & \cellcolor{red}0.030 & \cellcolor{red}0.031 & \cellcolor{red}0.050 & \cellcolor{red}0.039
\\
Mip-Splatting\cite{yu2023mip}
& \cellcolor{orange}36.50 & \cellcolor{orange}30.72 & \cellcolor{orange}27.81 & \cellcolor{orange}26.51 & \cellcolor{orange}30.39
& \cellcolor{orange}0.986 & \cellcolor{orange}0.959 & \cellcolor{orange}0.920 & \cellcolor{orange}0.893 & \cellcolor{orange}0.939
& \cellcolor{orange}0.015 & \cellcolor{orange}0.048 & \cellcolor{orange}0.099 & \cellcolor{orange}0.130 & \cellcolor{orange}0.073
\\
\hline
$SA\text{-}GS_{fil} \text{(ours)}$
& \cellcolor{yellow}35.74 & \cellcolor{yellow}30.38 & \cellcolor{yellow}27.63 & \cellcolor{yellow}26.36 & \cellcolor{yellow}30.03
& \cellcolor{yellow}0.984 & \cellcolor{yellow}0.953 & \cellcolor{yellow}0.912 & \cellcolor{yellow}0.885 & \cellcolor{yellow}0.933
& \cellcolor{yellow}0.016 & \cellcolor{yellow}0.059 & \cellcolor{yellow}0.111 & \cellcolor{yellow}0.141 & \cellcolor{yellow}0.082
\\
\end{tabular}
    }
    \vspace{0.5em}
    \caption{
         \textbf{Single-scale training and Multi-scale testing on the Blender dataset\cite{mildenhall2020nerf} for zoom-in effect.} We use the experiment and model naming with the Mip-NeRF 360\cite{barron2022mip} experiment (of Table ~\ref{tab:avg_single360_zoom_in_results}). Our methods yields comparable results with $Mip\text{-}Splatting$\cite{yu2023mip}. $SA\text{-}GS_{fil}$ achieves this performance while being training-free.
    }
    \vspace{-0.5em}
    \label{tab:avg_singleblender_zoom_in_results}
\end{table*}

\section{Conclusion}
We present SA-GS, a training-free framework that can seamlessly integrate with 3DGS\cite{kerbl20233dgs} to enhance its anti-aliasing ability at arbitrary rendering frequencies. Specifically, we propose a 2D scale-adaptive filter, which maintains the 2D Gaussian projection scale's consistency under different rendering settings. In addition, we employ conventional anti-aliasing techniques, super-sampling, and integration to significantly reduce image aliasing at lower sampling rates. SA-GS demonstrates superior or comparable performance to the state-of-the-art, as extensive validation is performed on both bounded and unbounded scenarios.\\
\textbf{Limitations.} Our method has no computational burden when zooming in, but when zooming out, the application of integration and super-sampling methods increases the rendering time. Due to shared memory, the elapsed time for super-sampling is comparable to that of integration, making it 15\%$\sim$20\% slower than the vanilla 3DGS\cite{kerbl20233dgs}. However, integration can still be optimized(approximation calculations or table lookups), leading to further speedups. Overall, our approach receives a significant anti-aliasing performance boost with minimal trade-offs.

%
%
\bibliographystyle{splncs04}
\bibliography{main}

\clearpage
\renewcommand{\thesection}{\Alph{section}}
\setcounter{figure}{9}
\setcounter{table}{6}
\title{SA-GS: Scale-Adaptive Gaussian Splatting for Training-Free Anti-Aliasing}
\author{\large \textbf{Supplementary Material}}
\institute{}
\maketitle

In this supplementary material, we first present the implementation details of the integration in \ref{A}. Next, we proof the theoretical upper bound for the rotational error in \ref{B.1} and present the results of numerical experiments in \ref{B.2}. Additionally, we present ablation studies of SA-GS in \ref{C}. Finally, we report additional quantitative and qualitative results in \ref{D}.

\section{Implementation Details of Integration}
\label{A}

As stated in the main text, we simplify the integral operation by \textit{rotating the pixel}. In the concrete implementation, the rotation is simulated by projection. Denote the two normalised eigenvectors of the 2D Gaussian distribution as $\Vec{v}_{long}$ and $\Vec{v}_{short}$, corresponding to larger and smaller eigenvalues respectively. We project the four corner points of the pixel in the direction of $\Vec{v}_{long}$ and $\Vec{v}_{short}$ using dot product:

\begin{equation}
    \begin{aligned}
       x_{min} &= min(P_{lu,lb,ru,rb}\cdot \Vec{v}_{long})\\
       x_{max} &= max(P_{lu,lb,ru,rb}\cdot \Vec{v}_{long})\\
       y_{min} &= min(P_{lu,lb,ru,rb}\cdot \Vec{v}_{short})\\
       y_{max} &= max(P_{lu,lb,ru,rb}\cdot \Vec{v}_{short})
    \end{aligned}
\end{equation}

where $P_{lu,lb,ru,rb}$ are the coordinates of the four corner points of the pixel, specifically the left upper, left lower, right upper and right lower. $x_{max}-x_{min}$ or $y_{max}-y_{min}$ is equivalent to the side length of the rotated pixel.

\begin{figure}[t]
  \centering
  \includegraphics[height=3.5cm]{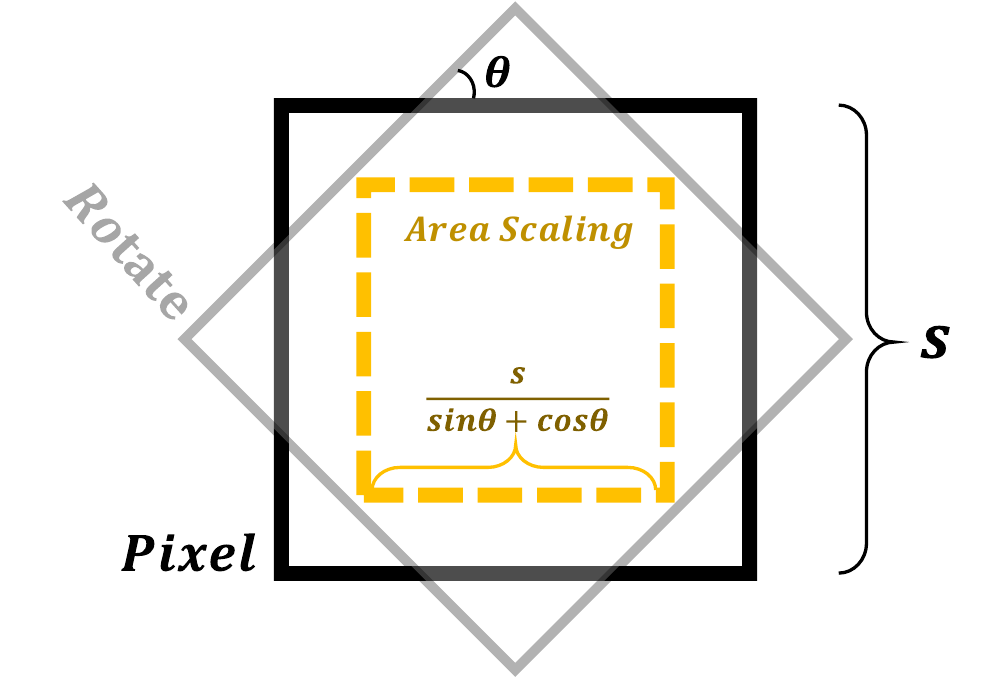}
  \caption{\textbf{Area scaling when rotating pixels.} In integration method, the pixel area is scaled before projection to ensure that the projected (rotated) pixel area is equal to the original pixel area. $\theta$ is the rotation angle of the pixel.}
  \label{fig:pixel rotate}
\end{figure}

However, when rotating pixels in the above manner, their edge lengths always increase, resulting in an area that is larger than the correct range. On the other hand, restricting the area of the rotated pixel to be inside the original pixel causes a loss of integral area. To balance this issue, we scale the pixel area before projection to ensure that the rotated pixel area is equal to the original pixel area. Specifically, we multiply the original pixel edge lengths by the $\frac{1}{sin\theta+cos\theta}$, as illustrated in Fig. \ref{fig:pixel rotate}.

\section{Analysis of Rotational Errors}
\label{B}
Although rotating the pixel simplifies the integral calculation, it inevitably introduces an error, even if the area of the pixel after rotation is equal to the original. We prove that for any pixel close enough to the center of the Gaussian to be affected during $\alpha$-blending, there exists a theoretical upper bound for the error. Additionally, We verify through numerical experimentation that this is empirically a good approximation.

\subsection{Theoretical Upper Bound}
\label{B.1}
\subsubsection{Normalization and Rotation} Let the center of the pixel be at coordinates $(x_c, y_c)$, and the side length of the pixel be $l$. Let the pixel have a counterclockwise tilt of $\theta$ with respect to the standard x-axis. The first problem we need to solve would be to eliminate the tilt in order to simplify the double integral. In order to do so, we would need to apply a rotation transformation to the Bivariate Gaussian Distribution without changing its main form. Hence we first normalize it into a Bivariate Normal Distribution via scaling. 

We construct $x^*,y^*=\frac{x}{\sigma_x},\frac{y}{\sigma_y}$ such that we have the Normal Distribution $g_x^*(x^*)=\frac{\exp(-\frac12x^{*2})}{\sqrt{2\pi}}=\sigma_xg_x(x)$ ($g_y^*$ is analogous). Hence we have: 

\begin{equation}
\label{scale}
f^*(x^*,y^*)=\sigma_x\sigma_yg_x(x)g_y(y)=\sigma_x\sigma_yf(x,y)
\end{equation}

The rotational symmetry of the Bivariate Normal Distribution now allows us to rotate the pixel clockwise by $\theta$ with respect to the origin without changing the integral (this does not affect the integral). Now the new pixel region, $P^*$, is a parallelogram with top and bottom edges parallel to the x-axis. Let the four corners (labeled in order analogously to quadrants) be labeled respectively $P_1(x_1, y_1)$, $P_2(x_2, y_1)$, $P_3(x_3, y_3)$, $P_4(x_4, y_3)$, and the slope of the edges $P_1P_4$ and $P_2P_3$ be $k$. WLOG let $y_1\ge0$ (otherwise conduct a reflection across the x-axis).

\subsubsection{Double Integration} In order to find bounds for $\iint\limits_{P^*}f^*(x^*, y^*)dx^*dy^*$, if we keep the current boundary for $P^*$, we will need to integrate $\Phi g$, which would be unrepresentable. In order to find the theoretical range of the error, we instead try to fix the sliced Gaussian distribution by fixing $y^*$. Combining with (\ref{scale}), we readily have the following bounds:

\begin{align}
\iint\limits_{P}f(x,y)dxdy&>\frac1{\sigma_x\sigma_y}\int_{y_3}^{y_1}\int_{\frac{y^*-y_1}k+x_2}^{\frac{y-y_1}k+x_1}f^*(x,\max\{0, y_3\})dxdy \\
\iint\limits_{P}f(x,y)dxdy&<\frac1{\sigma_x\sigma_y}\int_{y_3}^{y_1}\int_{\frac{y^*-y_1}k+x_2}^{\frac{y-y_1}k+x_1}f^*(x,\max\{y_1, |y_3|\})dxdy
\end{align}

\begin{figure}[t]
\centering
\includegraphics[width=.8\textwidth]{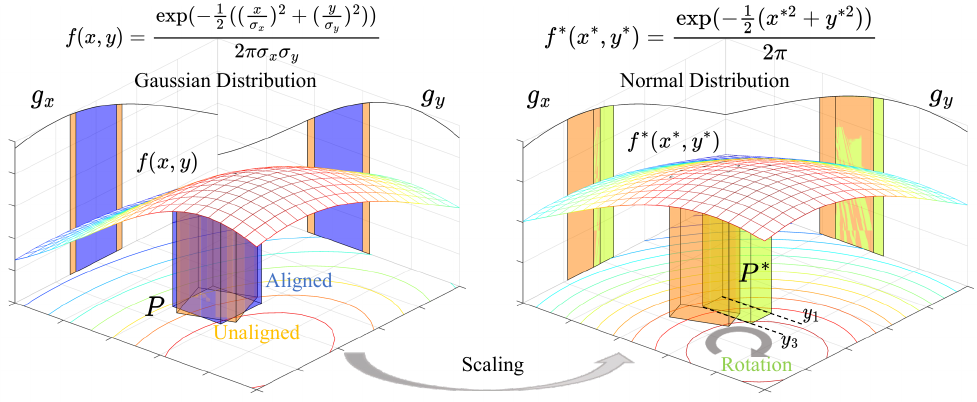}
\caption{\textbf{Visual demonstration of our theoretical analysis.} We scale the general Gaussian distribution to a standard normal distribution to estimate an upper bound on the error between the rotated pixel and the original pixel.}
\label{theory_vis}
\end{figure}

Fig. \ref{theory_vis} presents a visualisation of the theoretical analysis. Obviously, when the coordinates on the pixel are bounded by a constant times the respective standard deviations, coordinates of $P_i$ will also be bounded by the constant. Further, $k$ is only affected by $\theta$, $\sigma_x$, and $\sigma_y$. Hence the original double integral is bounded in our approximation, and so is its error.

\subsection{Numerical Experiments}
\label{B.2}
We verify the above theoretical analysis by numerical experiments. Adopting the notation established in \ref{B.1}, we denote $(x_c, y_c)$ as the coordinates of the pixel centroid, $l$ as the pixel side length, $\theta$ as the angle defining the counterclockwise rotation of the pixel, and $\sigma_x$ and $\sigma_y$ as the standard deviations corresponding to the directions of the two principal eigenvectors of the Gaussian distribution, respectively.

\begin{figure}
  \centering
  \includegraphics[width=\textwidth]{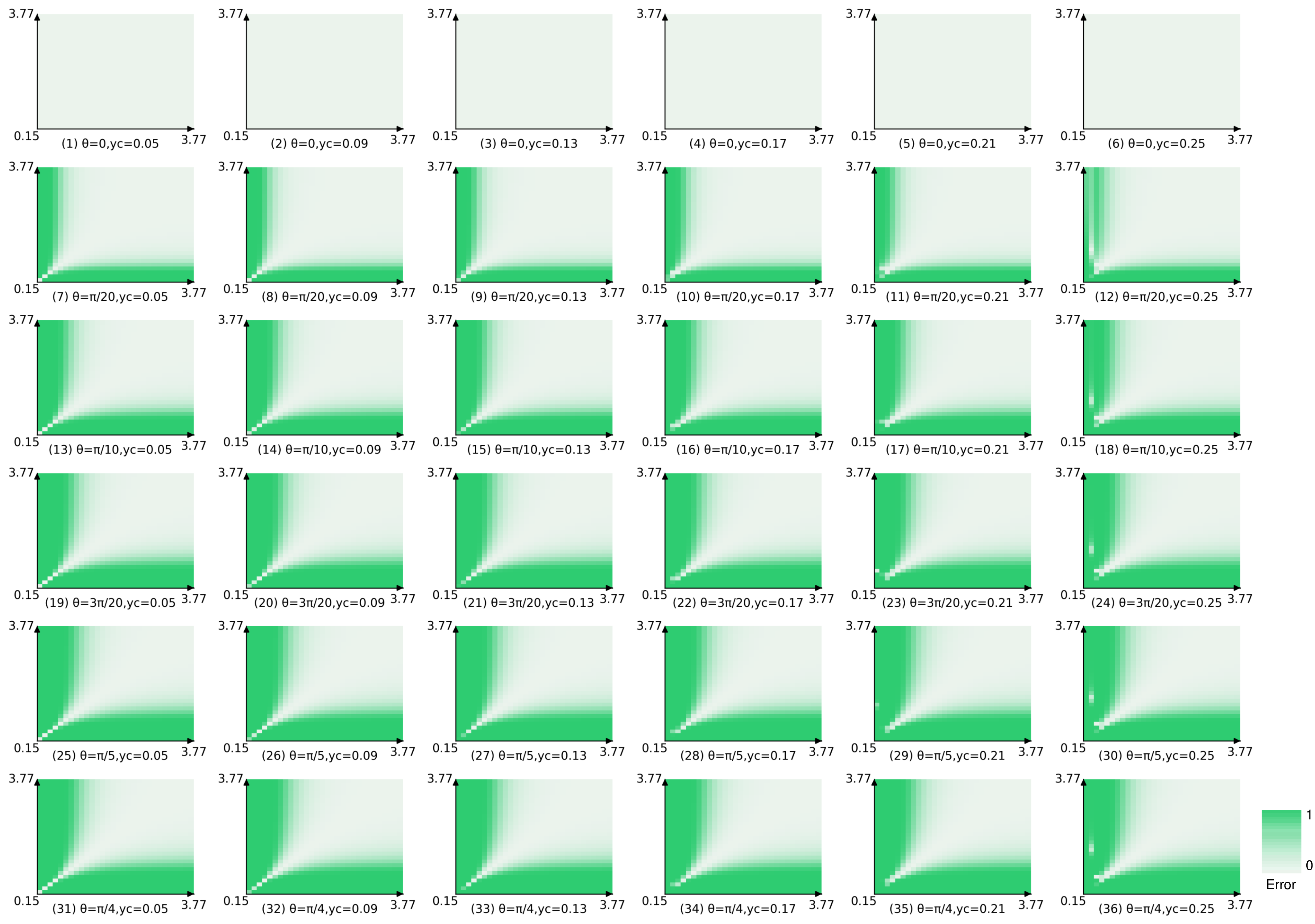}
  \caption{\textbf{Numerical Experimental Results of Integration Error.} we convert all the errors to the (0,1) range after transformation and widened the differences in the region close to 0. The average relative error is \textbf{0.51\%}, verifying that our method is a good estimation.}
  \label{fig:Numerical Experiments}
\end{figure}

We adopted the framework of parametric sensitivity analysis for our experimental setup, by fixing certain parameters while sampling within reasonable ranges for the others. This approach aims to quantify the differences between the original pixel integrals and their counterparts after rotation. Specifically, we set $l=1$ and $x_c=0$, allowing $\theta$ and $y_c$ to uniformly sample six values from their respective intervals $[0, \frac{\pi}{4}]$ and $[0.05, 0.25]$, thereby generating 36 sub-tables, as depicted in Fig. \ref{fig:Numerical Experiments}. In each sub-table, the parameters $\sigma_x$ and $\sigma_y$ delineate the horizontal and vertical axes, correspondingly, with both parameters uniformly sampling 30 values from the interval $[0.15, 3.77]$. This interval encompasses the core portion of the Gaussian distribution represented in the tile.

The final numerical experiment give an average relative error of \textbf{0.51\%}. Since most of the errors are 0 or close to 0, for ease of visualisation, we convert all the errors to the (0,1) range and widen the differences in the region close to 0 by $y=\frac{1}{1+e^{-800x}}$, as shown in Fig. \ref{fig:Numerical Experiments}. It can be seen that the error increases as the anisotropy of the Gaussian distribution becomes more pronounced, and that the error range increases as $\theta$ increases. However, the overall error values calculated are small, confirming that our method is a good estimation.

\section{Ablation}
\label{C}
In this section, we evaluate the effectiveness of 2D scale-adaptive filter(\ref{C.1}) and anti-aliasing methods(\ref{C.2}). Additionally, we present corresponding qualitative and quantitative results.

\subsection{Effectiveness of the 2D Scale-adaptive Filter}
\label{C.1}
To evaluate the effectiveness of the 2D Scale-adaptive filter, we perform ablation studies with single-scale training and multi-scale testing(zoom-out and zoom-in) on both the Mip-NeRF 360 dataset and the Blender dataset. The quantitative results are presented in Table \ref{tab:ablation_360_zoom_out_results}, Table \ref{tab:ablation_360_zoom_in_results}, Table \ref{tab:ablation_blender_zoom_out_results}, and Table \ref{tab:ablation_blender_zoom_in_results}.

\begin{figure}
\centering
\includegraphics[width=\textwidth]{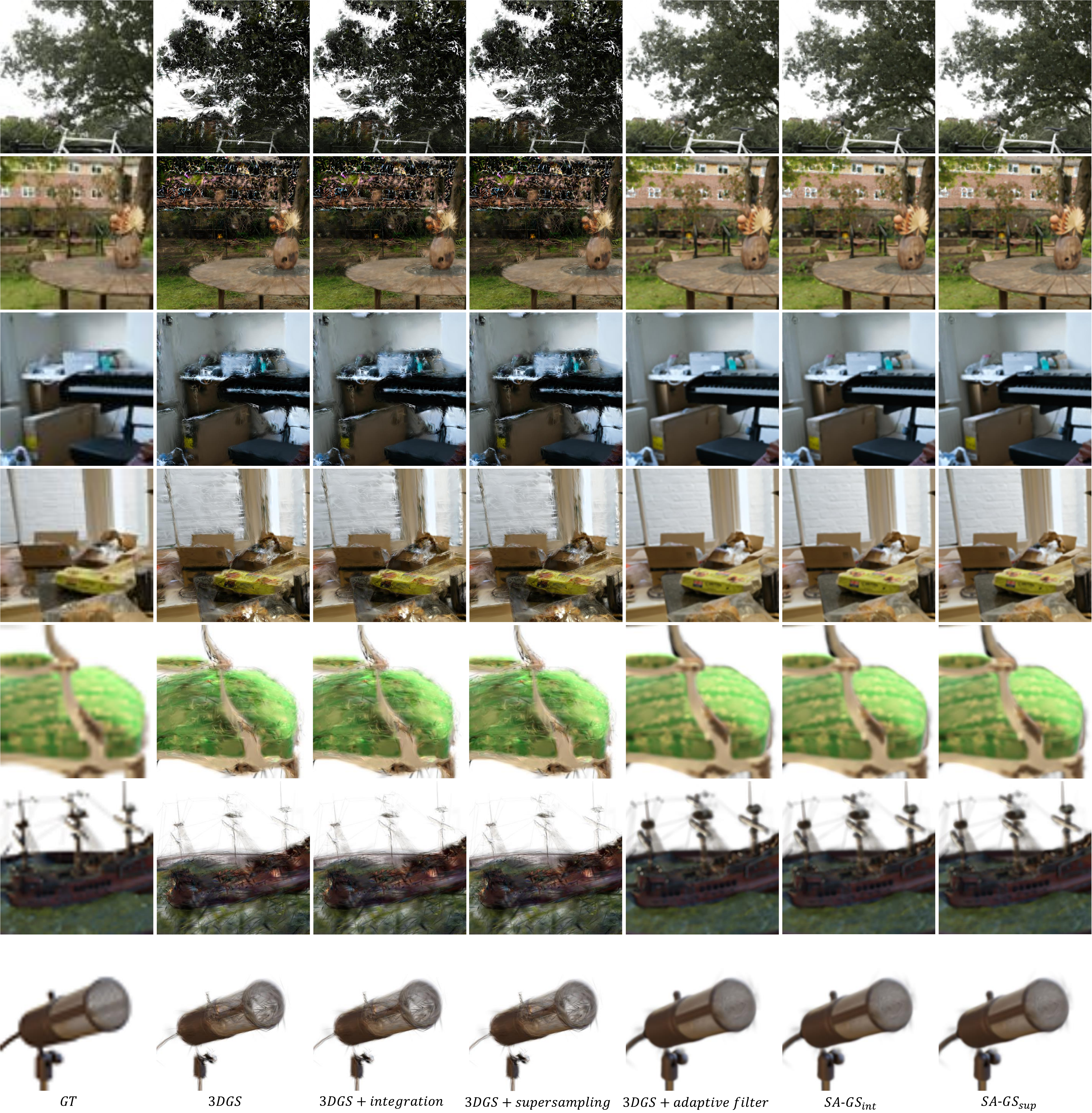}
\caption{\textbf{Single-scale Training and Multi-scale Testing for Zoom-in Effect.} All method are trained on smallest resolution($\nicefrac{1}{8}\times$) and evaluated on full resolution($1\times$) to mimic zoom-in case. 3DGS suffers from bloat or erosion artefacts at different rendering frequencies, which can exacerbate the aliasing effect. Our 2D scale-adaptive filter maintains the scale consistency of the Gaussian across different rendering settings. Note that the integration and super-sampling methods are only designed for the zoom-out case, so in the zoom-in case they maintain comparable effects to the addition of a 2D scale-adaptive filter.}
\label{fig:ablation_zoomin}
\end{figure}

Due to the scale consistency across rendering settings brought about by the 2D scale-adaptive filter, we get a very noticeable performance improvement over 3DGS in both zoom-out and zoom-in scenarios. 3DGS expands or shrinks at different rendering frequencies, thus exacerbating the aliasing effect, as illustrated in Fig. \ref{fig:ablation_zoomout} and Fig. \ref{fig:ablation_zoomin}.

\subsection{Effectiveness of the Anti-aliasing Methods}
\label{C.2}
To evaluate the effectiveness of the anti-aliasing methods(integration and super-sampling), we perform ablation studies with single-scale training and multi-scale testing for zoom-out effect on both the Mip-NeRF 360 dataset and the Blender dataset. The quantitative results are presented in Table \ref{tab:ablation_360_zoom_out_results} and Table \ref{tab:ablation_blender_zoom_out_results}. Note that the integration and super-sampling methods are intended solely for decreasing rendering frequency. Therefore, we do not focus on analysing their performances in the zoom-in case. Table \ref{tab:ablation_360_zoom_in_results} and Table \ref{tab:ablation_blender_zoom_in_results} demonstrate that they perform comparably with 3DGS.

\begin{figure}
\centering
\includegraphics[width=\textwidth]{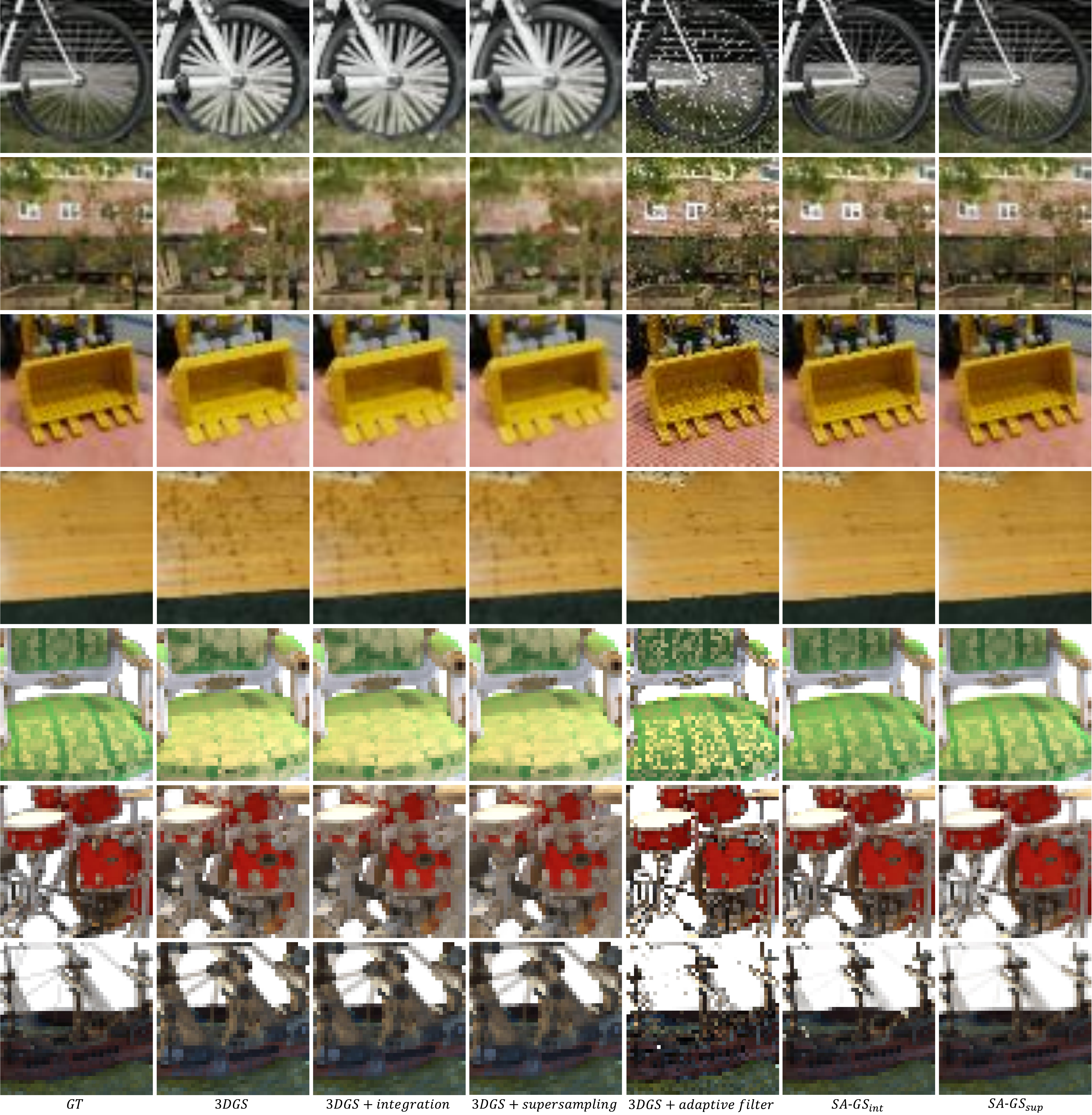}
\caption{\textbf{Single-scale Training and Multi-scale Testing for Zoom-out Effect.} All method are trained on full resolution($1\times$) and evaluated on smallest resolution($\nicefrac{1}{8}\times$) to mimic zoom-out case. 3DGS suffers from bloat or erosion artefacts at different rendering frequencies, which can exacerbate the aliasing effect. Our 2D scale-adaptive filter maintains the scale consistency of the Gaussian across different rendering settings. Additionally, our integration and super-sampling methods further enhance the anti-aliasing ability of Gaussian scenes. It is important to note that the integration and super-sampling methods are only effective in combination with the 2D scale-adaptive filter.}
\label{fig:ablation_zoomout}
\end{figure}

The integration and super-sampling methods are ineffective when the 2D scale-adaptive filter fails due to scale inconsistency in the 3DGS. However, when the 2D scale-adaptive filter is operational, these methods can further enhance the anti-aliasing ability of the scene, as illustrated in Fig. \ref{fig:ablation_zoomout}. In summary, we conclude that 3DGS does not provide a more robust representation of the scene using conventional anti-aliasing methods, but our 2D scale-adaptive filter completely removes this limitation.

\begin{table*}
    \renewcommand{\tabcolsep}{1pt}
    \small
    \centering
    \resizebox{\linewidth}{!}{
        \begin{tabular}{l|ccccp{0.9cm}<{\centering}|ccccp{0.9cm}<{\centering}|ccccp{0.9cm}<{\centering}}
        & \multicolumn{5}{c|}{PSNR $\uparrow$} & \multicolumn{5}{c|}{SSIM $\uparrow$} & \multicolumn{5}{c}{LPIPS$\downarrow$}\\
              & 1 Res. & $\nicefrac{1}{2}$ Res. & $\nicefrac{1}{4}$ Res. & $\nicefrac{1}{8}$ Res. & Avg. & 1 Res. & $\nicefrac{1}{2}$ Res. & $\nicefrac{1}{4}$ Res. & $\nicefrac{1}{8}$ Res. & Avg. & 1 Res. & $\nicefrac{1}{2}$ Res. & $\nicefrac{1}{4}$ Res. & $\nicefrac{1}{8}$ Res & Avg. \\ \hline

3DGS
& \cellcolor{red}29.26 & 26.84 & 22.16 & 19.63 & 24.47
& \cellcolor{red}0.877 & 0.863 & 0.726 & 0.612 & 0.769
& \cellcolor{red}0.185 & 0.148	& 0.198 & 0.223 & 0.189
\\
3DGS+Integration
& 29.14 & 26.44 & 22.22 & 19.32 & 24.28
& 0.873	& 0.850	& 0.726	& 0.588	& 0.759
& 0.188	& 0.158	& 0.197	& 0.238	& 0.196
\\
3DGS+Super-sampling
& \cellcolor{red}29.26	& 26.88	& 22.79	& 19.91	& 24.71
& 0.876	& 0.861	& 0.754	& 0.633	& 0.781
& 0.186	& 0.153	& 0.188	& 0.223	& 0.188
\\
3DGS+Adaptive Filter
& \cellcolor{red}29.26 & \cellcolor{yellow}29.80 & \cellcolor{yellow}28.26 & \cellcolor{yellow}25.58 & \cellcolor{yellow}28.23
& \cellcolor{red}0.877 & \cellcolor{yellow}0.901 & \cellcolor{yellow}0.875 & \cellcolor{yellow}0.809 & \cellcolor{yellow}0.866
& \cellcolor{red}0.185 & \cellcolor{yellow}0.123 & \cellcolor{yellow}0.126 & \cellcolor{yellow}0.171 & \cellcolor{yellow}0.151
\\
Full Method($SA\text{-}GS_{int}$)
& 29.14	& \cellcolor{orange}{30.06} & \cellcolor{orange}30.13	& \cellcolor{orange}{28.81} & \cellcolor{orange}{29.53}
& 0.873	& \cellcolor{orange}{0.905} & \cellcolor{orange}0.921 & \cellcolor{orange}{0.919} & \cellcolor{orange}{0.904}
& 0.188 & \cellcolor{orange}{0.118} & \cellcolor{orange}0.086 & \cellcolor{orange}{0.078} & \cellcolor{orange}{0.118}
\\
Full Method($SA\text{-}GS_{sup}$)
& \cellcolor{red}{29.26} & \cellcolor{red}{30.45} & \cellcolor{red}{31.75} & \cellcolor{red}{32.53} & \cellcolor{red}{31.00}
& \cellcolor{orange}{0.876} & \cellcolor{red}{0.912} & \cellcolor{red}{0.938} & \cellcolor{red}{0.951} & \cellcolor{red}{0.919}
& \cellcolor{orange}{0.186} & \cellcolor{red}{0.114} & \cellcolor{red}{0.073} & \cellcolor{red}{0.053} & \cellcolor{red}{0.106}
\\
\end{tabular}
    }
    \vspace{0.5em}
    \caption{
        \textbf{Ablation studies for zoom-out effect on the Mip-NeRF 360 Dataset.} All methods are trained on the largest scale (1×) and evaluated across four scales (1×, \nicefrac{1}{2}×, \nicefrac{1}{4}×, and \nicefrac{1}{8}×). Our 2D scale-adaptive filter removes 3DGS bloat at low rendering frequencies. Additionally, our integration and super-sampling methods further enhance anti-aliasing ability, as illustrated in Fig. \ref{fig:ablation_zoomout}. It is important to note that the integration and super-sampling methods are only effective when the 2D scale-adaptive filter is active.
    }
    \vspace{-2em}
    \label{tab:ablation_360_zoom_out_results}
\end{table*}

\begin{table*}
    \renewcommand{\tabcolsep}{1pt}
    \small
    \centering
    \resizebox{\linewidth}{!}{
        \begin{tabular}{l|ccccp{0.9cm}<{\centering}|ccccp{0.9cm}<{\centering}|ccccp{0.9cm}<{\centering}}
        & \multicolumn{5}{c|}{PSNR $\uparrow$} & \multicolumn{5}{c|}{SSIM $\uparrow$} & \multicolumn{5}{c}{LPIPS$\downarrow$}\\
              & 1 Res. & 2 Res. & 4 Res. & 8 Res. & Avg. & 1 Res. & 2 Res. & 4 Res. & 8 Res. & Avg. & 1 Res. & 2 Res. & 4 Res. & 8 Res & Avg. \\ \hline

3DGS
& \cellcolor{red}33.96	& 22.47 & 18.69 & \cellcolor{yellow}17.32 & \cellcolor{orange}23.11
& \cellcolor{red}0.974	& 0.747 & 0.514 & 0.460 & 0.674
& \cellcolor{red}0.028 & \cellcolor{orange}0.204	& 0.410 & 0.504 & 0.286
\\
3DGS+Integration
& 32.57	& \cellcolor{orange}22.67	& \cellcolor{yellow}18.74	& 17.30	& 22.82
& 0.962	& \cellcolor{orange}0.754	& \cellcolor{orange}0.520	& \cellcolor{yellow}0.463	& \cellcolor{yellow}0.675
& 0.040	& \cellcolor{orange}0.204	& \cellcolor{orange}0.407	& \cellcolor{yellow}0.502	& \cellcolor{orange}0.288
\\
3DGS+Super-sampling
& \cellcolor{orange}33.05	& \cellcolor{yellow}22.65	& \cellcolor{orange}18.77	& \cellcolor{orange}17.36	& \cellcolor{yellow}22.96
& \cellcolor{yellow}0.966	& \cellcolor{yellow}0.753	& \cellcolor{orange}0.520	& \cellcolor{orange}0.464	& \cellcolor{orange}0.676
& \cellcolor{orange}0.038	& 0.205	& \cellcolor{orange}0.407	& \cellcolor{orange}0.501	& \cellcolor{orange}0.288
\\
3DGS+Adaptive Filter
& \cellcolor{red}33.96	& \cellcolor{red}27.89	& \cellcolor{red}25.32	& \cellcolor{red}24.40	& \cellcolor{red}27.89
& \cellcolor{red}0.974	& \cellcolor{red}0.840	& \cellcolor{red}0.677	& \cellcolor{red}0.615	& \cellcolor{red}0.777
& \cellcolor{red}0.028	& \cellcolor{red}0.189	& \cellcolor{red}0.360	& \cellcolor{red}0.465	& \cellcolor{red}0.260
\\
\end{tabular}
    }
    \vspace{0.5em}
    \caption{
        \textbf{Ablation studies for zoom-in effect on the Mip-NeRF 360 Dataset.} All methods are trained on the \nicefrac{1}{8} scale (1×) and evaluated across four scales (1×, 2×, 4×, and 8×). Our 2D scale-adaptive filter eliminates erosion artefacts at high rendering frequencies, as illustrated in Fig. \ref{fig:ablation_zoomin}. Integration and super-sampling methods are not designed for this case, which are comparable to 3DGS.
    }
    \vspace{-2em}
    \label{tab:ablation_360_zoom_in_results}
\end{table*}

\begin{table*}
    \renewcommand{\tabcolsep}{1pt}
    \small
    \centering
    \resizebox{\linewidth}{!}{
        \begin{tabular}{l|ccccp{0.9cm}<{\centering}|ccccp{0.9cm}<{\centering}|ccccp{0.9cm}<{\centering}}
        & \multicolumn{5}{c|}{PSNR $\uparrow$} & \multicolumn{5}{c|}{SSIM $\uparrow$} & \multicolumn{5}{c}{LPIPS$\downarrow$}\\
              & 1 Res. & $\nicefrac{1}{2}$ Res. & $\nicefrac{1}{4}$ Res. & $\nicefrac{1}{8}$ Res. & Avg. & 1 Res. & $\nicefrac{1}{2}$ Res. & $\nicefrac{1}{4}$ Res. & $\nicefrac{1}{8}$ Res. & Avg. & 1 Res. & $\nicefrac{1}{2}$ Res. & $\nicefrac{1}{4}$ Res. & $\nicefrac{1}{8}$ Res & Avg. \\ \hline

3DGS 
& \cellcolor{red}35.10 & 27.91 & 22.42 & 18.76 & 26.05
& \cellcolor{red}0.974 & 0.949 & 0.862 & 0.736 & 0.880
& \cellcolor{red}0.029 & 0.033 & 0.069 & 0.133 & 0.066
\\
3DGS+Integration
& 34.49 & 28.06 & 22.78 & 19.12 & 26.11
& 0.972 & 0.948 & 0.867 & 0.745 & 0.883
& 0.032 & 0.036 & 0.072 & 0.135 & 0.069
\\
3DGS+Super-sampling
& 34.27 & 27.10 & 21.93 & 18.39 & 25.42
& 0.972 & 0.939 & 0.848 & 0.719 & 0.869
& 0.032 & 0.038 & 0.078 & 0.146 & 0.074
\\
3DGS+Adaptive Filter
& \cellcolor{orange}34.60 & \cellcolor{yellow}34.33 & \cellcolor{orange}31.02 & \cellcolor{orange}27.59 & \cellcolor{orange}31.89
& \cellcolor{orange}0.973 & \cellcolor{yellow}0.977 & \cellcolor{yellow}0.968 & \cellcolor{orange}0.947 & \cellcolor{orange}0.966
& \cellcolor{orange}0.031 & \cellcolor{yellow}0.022 & \cellcolor{yellow}0.036 & \cellcolor{yellow}0.067 & \cellcolor{yellow}0.039
\\
Full Method($SA\text{-}GS_{int}$)
& 34.35 & \cellcolor{orange}34.39 & \cellcolor{yellow}30.99 & \cellcolor{yellow}26.89 & \cellcolor{yellow}31.65
& \cellcolor{yellow}0.972 & \cellcolor{orange}0.978 & \cellcolor{orange}0.971 & \cellcolor{yellow}0.940 & \cellcolor{yellow}0.965
& \cellcolor{yellow}0.032 & \cellcolor{orange}0.020 & \cellcolor{orange}0.023 & \cellcolor{orange}0.039 & \cellcolor{orange}0.029
\\
Full Method($SA\text{-}GS_{sup}$)
& \cellcolor{yellow}34.49 & \cellcolor{red}36.58 & \cellcolor{red}37.50 & \cellcolor{red}35.64 & \cellcolor{red}36.06
& \cellcolor{yellow}0.972 & \cellcolor{red}0.980 & \cellcolor{red}0.985 & \cellcolor{red}0.985 & \cellcolor{red}0.981
& \cellcolor{yellow}0.032 & \cellcolor{red}0.018 & \cellcolor{red}0.014 & \cellcolor{red}0.013 & \cellcolor{red}0.019
\end{tabular}
    }
    \vspace{0.5em}
    \caption{
        \textbf{Ablation studies for zoom-out effect on the Blender Dataset.} All methods are trained on the largest scale (1×) and evaluated across four scales (1×, \nicefrac{1}{2}×, \nicefrac{1}{4}×, and \nicefrac{1}{8}×). Our 2D scale-adaptive filter removes 3DGS bloat at low rendering frequencies. Additionally, our integration and super-sampling methods further enhance anti-aliasing ability, as illustrated in Fig. \ref{fig:ablation_zoomout}. It is important to note that the integration and super-sampling methods are only effective when the 2D scale-adaptive filter is active.
    }
    \vspace{-2em}
    \label{tab:ablation_blender_zoom_out_results}
\end{table*}

\begin{table*}
    \renewcommand{\tabcolsep}{1pt}
    \small
    \centering
    \resizebox{\linewidth}{!}{
        \begin{tabular}{l|ccccp{0.9cm}<{\centering}|ccccp{0.9cm}<{\centering}|ccccp{0.9cm}<{\centering}}
        & \multicolumn{5}{c|}{PSNR $\uparrow$} & \multicolumn{5}{c|}{SSIM $\uparrow$} & \multicolumn{5}{c}{LPIPS$\downarrow$}\\
              & 1 Res. & 2 Res. & 4 Res. & 8 Res. & Avg. & 1 Res. & 2 Res. & 4 Res. & 8 Res. & Avg. & 1 Res. & 2 Res. & 4 Res. & 8 Res & Avg. \\ \hline

3DGS
& \cellcolor{red}36.97 & 24.33 & 21.01 & 19.63 & \cellcolor{orange}25.44
& \cellcolor{red}0.988 & \cellcolor{orange}0.886 & \cellcolor{orange}0.820 & \cellcolor{orange}0.821 & \cellcolor{orange}0.879
& \cellcolor{red}0.013 & \cellcolor{orange}0.065 & \cellcolor{orange}0.130 & \cellcolor{orange}0.159 & \cellcolor{orange}0.092
\\
3DGS+Integration
& \cellcolor{yellow}34.12 & \cellcolor{orange}24.48 & \cellcolor{orange}21.09 & \cellcolor{orange}19.67 & \cellcolor{yellow}24.84
& \cellcolor{yellow}0.980 & \cellcolor{orange}0.886 & \cellcolor{yellow}0.819 & \cellcolor{yellow}0.820 & \cellcolor{yellow}0.876
& 0.025 & 0.072 & 0.133 & \cellcolor{yellow}0.161 & 0.098
\\
3DGS+Super-sampling
& 34.01 & \cellcolor{yellow}24.39 & \cellcolor{yellow}21.02 & \cellcolor{yellow}19.63 & 24.76
& \cellcolor{yellow}0.980 & 0.884 & 0.818 & \cellcolor{yellow}0.820 & \cellcolor{yellow}0.876
& \cellcolor{yellow}0.022 & \cellcolor{yellow}0.069 & \cellcolor{yellow}0.132 & \cellcolor{yellow}0.161 & \cellcolor{yellow}0.096
\\
3DGS+Adaptive Filter
& \cellcolor{orange}35.74 & \cellcolor{red}30.38 & \cellcolor{red}27.63 & \cellcolor{red}26.36 & \cellcolor{red}30.03
& \cellcolor{orange}0.984 & \cellcolor{red}0.953 & \cellcolor{red}0.912 & \cellcolor{red}0.885 & \cellcolor{red}0.933
& \cellcolor{orange}0.016 & \cellcolor{red}0.059 & \cellcolor{red}0.111 & \cellcolor{red}0.141 & \cellcolor{red}0.082
\\
\end{tabular}
    }
    \vspace{0.5em}
    \caption{
        \textbf{Ablation studies for zoom-in effect on the Blender Dataset.} All methods are trained on the \nicefrac{1}{8} scale (1×) and evaluated across four scales (1×, 2×, 4×, and 8×). Our 2D scale-adaptive filter eliminates erosion artefacts at high rendering frequencies, as illustrated in Fig. \ref{fig:ablation_zoomin}. Integration and super-sampling methods are not designed for this case, which are comparable to 3DGS.
    }
    \vspace{-2em}
    \label{tab:ablation_blender_zoom_in_results}
\end{table*}

\section{Additional Results}
\label{D}
In this section, we provide more qualitative and quantitative results on the Mip-NeRF 360 dataset(\ref{D.1}) and the Blender dataset(\ref{D.2}).
\subsection{Mip-NeRF 360 Dataset}
\label{D.1}
\begin{figure}
\centering
\includegraphics[width=\textwidth]{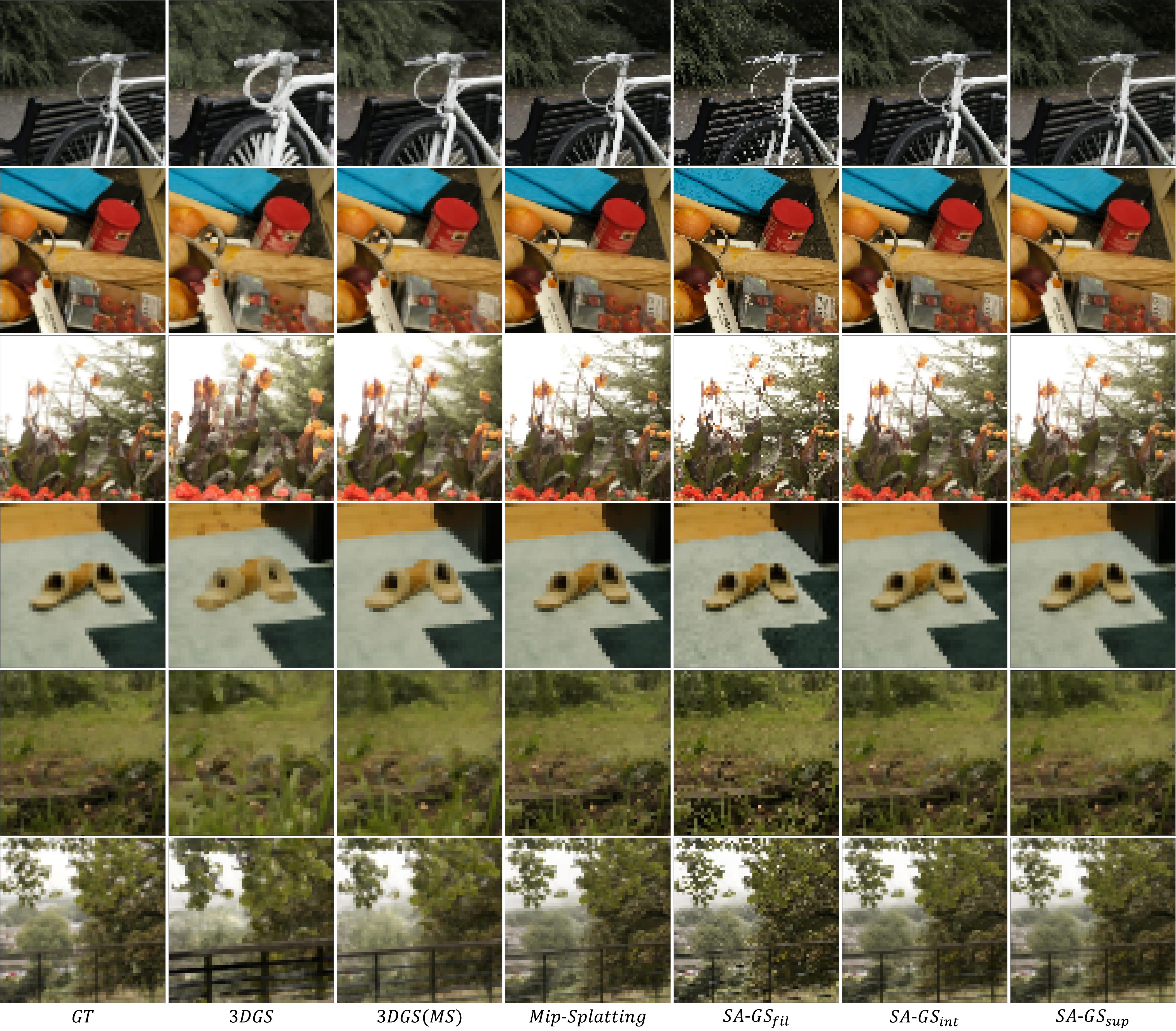}
\caption{\textbf{Single-scale Training and Multi-scale Testing on the Mip-NeRF 360 Dataset.} All method are trained on full resolution($1\times$) and evaluated on smallest resolution($\nicefrac{1}{8}\times$) to mimic zoom-out case. Our 2D scale-adaptive filter maintains the consistency of the Gaussian across different rendering settings. $SA\text{-}GS_{int}$ achieves results comparable to Mip-Splatting, while $SA\text{-}GS_{sup}$ surpasses Mip-Splatting, resulting in optimal performance for this setting.}
\label{fig:360_zoomout_add}
\end{figure}

\begin{figure}
\centering
\includegraphics[width=\textwidth]{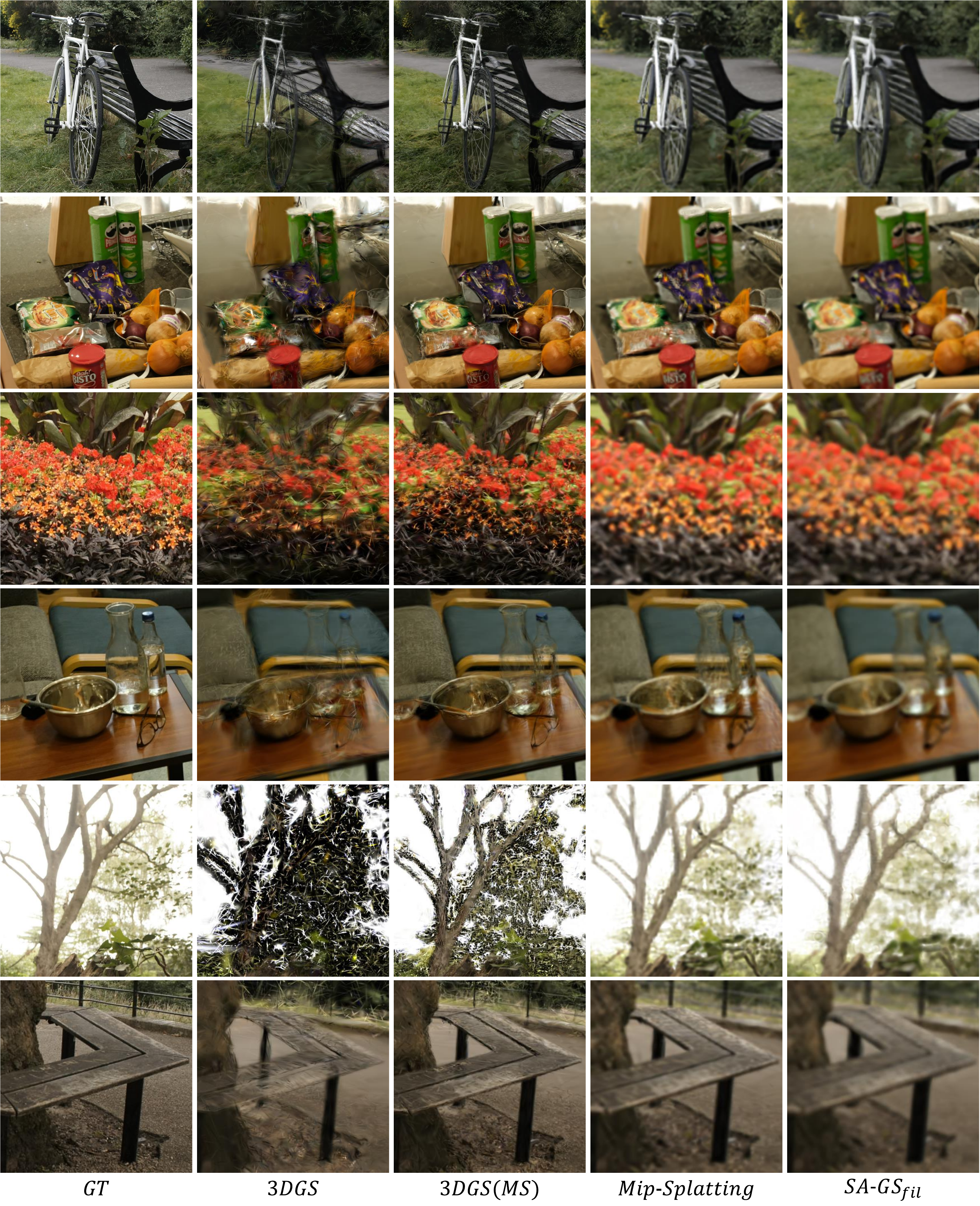}
\caption{\textbf{Single-scale Training and Multi-scale Testing on the Mip-NeRF 360 Dataset.} All method are trained on smallest resolution($\nicefrac{1}{8}\times$) and evaluated on full resolution($1\times$) to mimic zoom-in case. $SA\text{-}GS_{fil}$ achieves performance comparable to Mip-Splatting. However, Our 2D scale-adaptive filter is training-free and does not add any computational burden.}
\label{fig:360_zoomin_add}
\end{figure}

We further evaluate the effect of our method on zoom-out and zoom-in settings for each scene of this dataset. The quantitative results with per-scene metrics can be found in Table \ref{table:per_scene_360_zoomout} and Table \ref{table:per_scene_360_zoomin}. Qualitative comparison with state-of-the-art methods are provided in Fig. \ref{fig:360_zoomout_add} and Fig. \ref{fig:360_zoomin_add}. Our method achieves superior or comparable performance compared to the state-of-the-art, while being training-free.

\subsection{Blender Dataset}
\label{D.2}
\begin{figure}
\centering
\includegraphics[width=\textwidth]{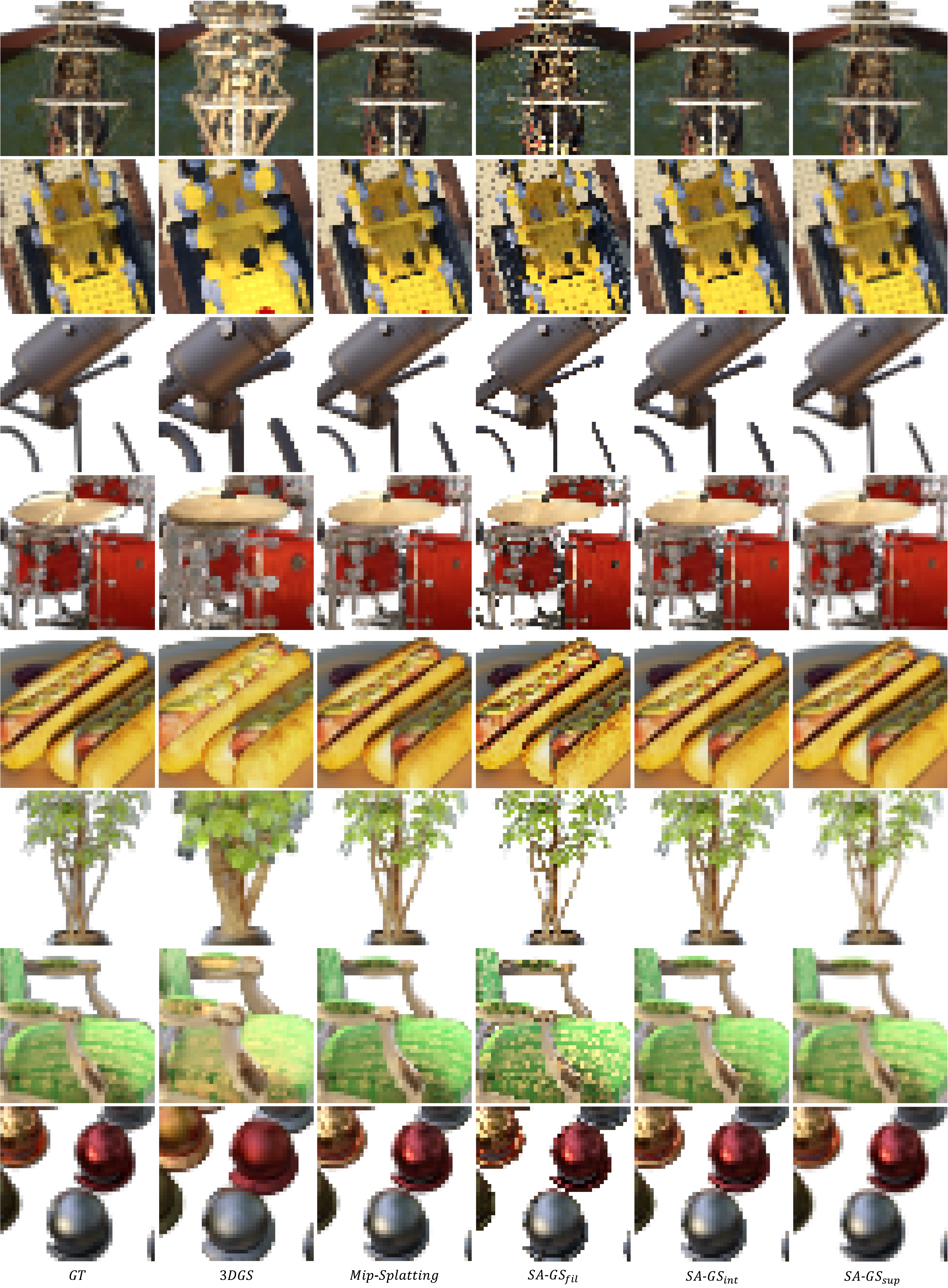}
\caption{\textbf{Single-scale Training and Multi-scale Testing on the Blender Dataset.} All method are trained on full resolution($1\times$) and evaluated on smallest resolution($\nicefrac{1}{8}\times$) to mimic zoom-out case. Our 2D scale-adaptive filter maintains the consistency of the Gaussian across different rendering settings. $SA\text{-}GS_{int}$ achieves results comparable to Mip-Splatting, while $SA\text{-}GS_{sup}$ surpasses Mip-Splatting, resulting in optimal performance for this setting.}
\label{fig:blender_zoomout_add}
\end{figure}

\begin{figure}
\centering
\includegraphics[width=0.8\textwidth]{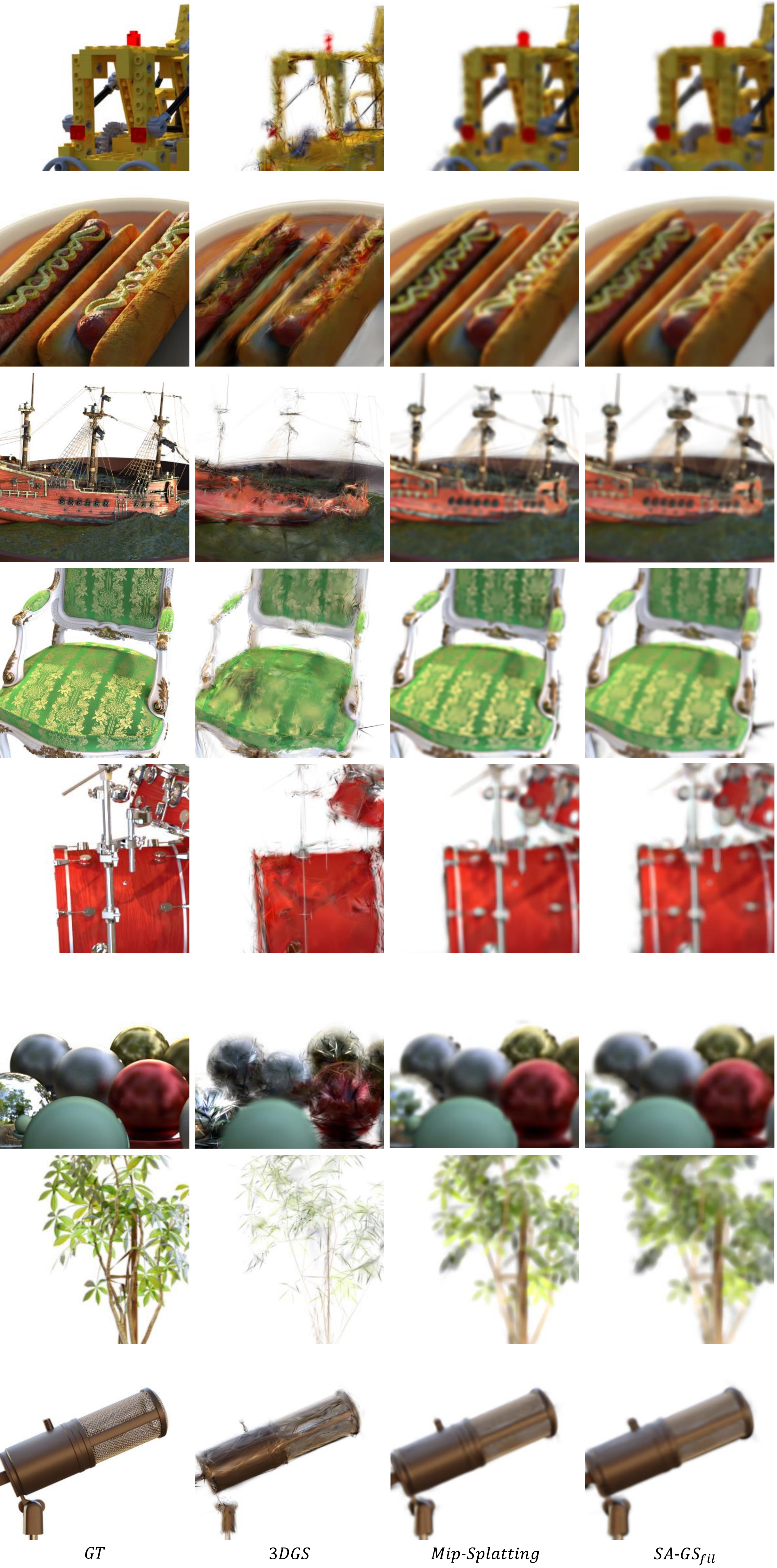}
\caption{\textbf{Single-scale Training and Multi-scale Testing on the Blender Dataset.} All method are trained on smallest resolution($\nicefrac{1}{8}\times$) and evaluated on full resolution($1\times$) to mimic zoom-in case. $SA\text{-}GS_{fil}$ achieves performance comparable to Mip-Splatting. However, Our 2D scale-adaptive filter is training-free and does not add any computational burden.}
\label{fig:blender_zoomin_add}
\end{figure}

We further evaluate the effect of our method on zoom-out and zoom-in settings for each scene of this dataset. The quantitative results with per-scene metrics can be found in Table \ref{table:per_scene_blender_zoomout} and Table \ref{table:per_scene_blender_zoomin}. Qualitative comparison with state-of-the-art methods are provided in Fig. \ref{fig:blender_zoomout_add} and Fig. \ref{fig:blender_zoomin_add}. Our method achieves superior or comparable performance compared to the state-of-the-art, while being training-free.

\begin{table}
\centering
\begin{subtable}{\linewidth}
\centering
\fontsize{7pt}{8.5pt}\selectfont
\begin{tabularx}{\linewidth}{l|>{\centering\arraybackslash}X>{\centering\arraybackslash}X>{\centering\arraybackslash}X>{\centering\arraybackslash}X>{\centering\arraybackslash}X>{\centering\arraybackslash}X>{\centering\arraybackslash}X>{\centering\arraybackslash}X>{\centering\arraybackslash}X|>{\centering\arraybackslash}X}
& \multicolumn{9}{c}{\textbf{PSNR $\uparrow$}}\\
& \textit{bonsai} & \textit{bicycle} & \textit{counter} & \textit{garden} & \textit{kitchen} & \textit{room} & \textit{stump} & \textit{flowers} & \textit{treehill} & \textit{Avg.}\\
\hline
3DGS
&27.06  &21.16  &26.13  &23.33  &26.69  &29.15  &25.00  &20.38  &21.37  &24.47
\\
3DGS(MS)
& 22.27 & \cellcolor{yellow}26.82 & 26.66 & 21.52 & 24.62 & 25.64 & 30.17 & 24.55 & 22.38 & 24.96
\\
Mip-Splatting
& 26.44 & \cellcolor{red}32.96 & \cellcolor{orange}30.42 & 25.63 & 30.54 & 32.88 & \cellcolor{red}34.01 & \cellcolor{red}30.86 & \cellcolor{orange}25.47 & \cellcolor{orange}29.91
\\
\hline
$SA\text{-}GS_{fil}(ours)$
&\cellcolor{yellow}31.74 &24.79 &\cellcolor{yellow}29.56 &\cellcolor{yellow}27.28	&\cellcolor{yellow}30.74 &\cellcolor{yellow}33.37 &28.98	&23.62 &23.98 &28.23
\\
$SA\text{-}GS_{int}(ours)$
&\cellcolor{orange}32.63 &26.21 &29.41 &\cellcolor{orange}30.22 &\cellcolor{orange}32.51 &\cellcolor{orange}33.85 &\cellcolor{yellow}30.58 &\cellcolor{yellow}25.25 &\cellcolor{yellow}25.14 &\cellcolor{yellow}29.53
\\
$SA\text{-}GS_{sup}(ours)$
&\cellcolor{red}34.13 &\cellcolor{orange}27.74 &\cellcolor{red}31.25 &\cellcolor{red}31.97 &\cellcolor{red}34.19 &\cellcolor{red}35.63 &\cellcolor{orange}31.89	&\cellcolor{orange}26.14 &\cellcolor{red}26.01 &\cellcolor{red}31.00
\end{tabularx}
\end{subtable}

\vspace{0.2cm}

\begin{subtable}{\linewidth}
\centering
\fontsize{7pt}{8.5pt}\selectfont
\begin{tabularx}{\linewidth}{l|>{\centering\arraybackslash}X>{\centering\arraybackslash}X>{\centering\arraybackslash}X>{\centering\arraybackslash}X>{\centering\arraybackslash}X>{\centering\arraybackslash}X>{\centering\arraybackslash}X>{\centering\arraybackslash}X>{\centering\arraybackslash}X|>{\centering\arraybackslash}X}
& \multicolumn{9}{c}{\textbf{SSIM $\uparrow$}}\\
& \textit{bonsai} & \textit{bicycle} & \textit{counter} & \textit{garden} & \textit{kitchen} & \textit{room} & \textit{stump} & \textit{flowers} & \textit{treehill} & \textit{Avg.}\\ 
\hline
3DGS
&0.872 &0.687 &0.847 &0.715 &0.835 &0.887 &0.744 &0.650 &0.687 &0.769
\\
3DGS(MS)
& 0.718 & \cellcolor{yellow}0.873 & 0.867 & 0.686 & 0.764 & 0.859 & \cellcolor{yellow}0.914 & 0.737 & 0.709 & 0.792
\\
Mip-Splatting
& 0.876 & \cellcolor{red}0.962 & \cellcolor{orange}0.942 & 0.820 & \cellcolor{yellow}0.934 & \cellcolor{orange}0.959 & \cellcolor{red}0.960 & \cellcolor{red}0.916 & \cellcolor{orange}0.836 & \cellcolor{orange}0.912
\\
\hline
$SA\text{-}GS_{fil}(ours)$
&\cellcolor{yellow}0.940 &0.813 &\cellcolor{yellow}0.921 &\cellcolor{yellow}0.835	&0.900 &0.946 &0.874	&0.774 &0.788 &0.866
\\
$SA\text{-}GS_{int}(ours)$
&\cellcolor{orange}0.960 &0.872 &0.906 &\cellcolor{orange}0.928 &\cellcolor{orange}0.957 &\cellcolor{orange}0.959 &0.911 &\cellcolor{yellow}0.814 &\cellcolor{yellow}0.832 &\cellcolor{yellow}0.904
\\
$SA\text{-}GS_{sup}(ours)$
&\cellcolor{red}0.966 &\cellcolor{orange}0.888 &\cellcolor{red}0.947 &\cellcolor{red}0.943	&\cellcolor{red}0.965 &\cellcolor{red}0.966 &\cellcolor{orange}0.923 &\cellcolor{orange}0.827 &\cellcolor{red}0.845 &\cellcolor{red}0.919
\end{tabularx}
\end{subtable}

\vspace{0.2cm}

\begin{subtable}{\linewidth}
\centering
\fontsize{7pt}{8.5pt}\selectfont
\begin{tabularx}{\linewidth}{l|>{\centering\arraybackslash}X>{\centering\arraybackslash}X>{\centering\arraybackslash}X>{\centering\arraybackslash}X>{\centering\arraybackslash}X>{\centering\arraybackslash}X>{\centering\arraybackslash}X>{\centering\arraybackslash}X>{\centering\arraybackslash}X|>{\centering\arraybackslash}X}
& \multicolumn{9}{c}{\textbf{LPIPS $\downarrow$}}\\
& \textit{bonsai} & \textit{bicycle} & \textit{counter} & \textit{garden} & \textit{kitchen} & \textit{room} & \textit{stump} & \textit{flowers} & \textit{treehill} & \textit{Avg.}\\
\hline
3DGS
&0.140 &0.239 &0.143  &0.169 &0.123 &0.134 &0.218 &0.269 &0.264 &0.189
\\
3DGS(MS)
& 0.228 & 0.154 & 0.148 & 0.251 & 0.183 & 0.140 & 0.131 & 0.234 & 0.261 & 0.192
\\
Mip-Splatting
& 0.136 & \cellcolor{red}0.086 & \cellcolor{orange}0.094 & 0.183 & \cellcolor{orange}0.061 & \cellcolor{red}0.060 & \cellcolor{red}0.088 & \cellcolor{red}0.118 & \cellcolor{orange}0.185 & \cellcolor{orange}0.112
\\
\hline
$SA\text{-}GS_{fil}(ours)$
&\cellcolor{yellow}0.109 &0.183 &0.120 &\cellcolor{yellow}0.135	&0.108 &0.112 &0.154	&0.215 &0.225 &0.151
\\
$SA\text{-}GS_{int}(ours)$
&\cellcolor{orange}0.088 &\cellcolor{yellow}0.139 &\cellcolor{yellow}0.119 &\cellcolor{orange}0.066 &\cellcolor{yellow}0.062 &\cellcolor{yellow}0.090 &\cellcolor{yellow}0.122 &\cellcolor{yellow}0.185 &\cellcolor{yellow}0.188 &\cellcolor{yellow}0.118
\\
$SA\text{-}GS_{sup}(ours)$
&\cellcolor{red}0.084 &\cellcolor{orange}0.125 &\cellcolor{red}0.091 &\cellcolor{red}0.054	&\cellcolor{red}0.055 &\cellcolor{orange}0.083 &\cellcolor{orange}0.110 &\cellcolor{orange}0.177 &\cellcolor{red}0.178 &\cellcolor{red}0.106
\end{tabularx}
\end{subtable}
\vspace{1em}
\caption{\textbf{Single-scale Training and Multi-scale testing on the Mip-NeRF 360 Dataset.} For each scene, we report the arithmetic mean of each metric averaged over the 4 scales($1\times$, $\nicefrac{1}{2}\times$, $\nicefrac{1}{4}\times$, $\nicefrac{1}{8}\times$) used in the dataset. (MS) means multi-scale training.}
\label{table:per_scene_360_zoomout}
\end{table}

\begin{table}
\centering
\begin{subtable}{\linewidth}
\centering
\fontsize{7pt}{8.5pt}\selectfont
\begin{tabularx}{\linewidth}{l|>{\centering\arraybackslash}X>{\centering\arraybackslash}X>{\centering\arraybackslash}X>{\centering\arraybackslash}X>{\centering\arraybackslash}X>{\centering\arraybackslash}X>{\centering\arraybackslash}X>{\centering\arraybackslash}X>{\centering\arraybackslash}X|>{\centering\arraybackslash}X}
& \multicolumn{9}{c}{\textbf{PSNR $\uparrow$}}\\
& \textit{bonsai} & \textit{bicycle} & \textit{counter} & \textit{garden} & \textit{kitchen} & \textit{room} & \textit{stump} & \textit{flowers} & \textit{treehill} & \textit{Avg.}\\
\hline
3DGS
&\cellcolor{yellow}24.52 &20.88 &24.64 &\cellcolor{yellow}22.57	&24.31 &\cellcolor{yellow}27.85 &22.45 &19.87 &20.89 &23.11
\\
3DGS(MS)
& 22.27 & \cellcolor{orange}26.82 & \cellcolor{yellow}26.66 & 21.52 & \cellcolor{yellow}24.62 & 25.64 & \cellcolor{orange}30.17 & \cellcolor{orange}24.55 & \cellcolor{yellow}22.38 & \cellcolor{yellow}24.96
\\
Mip-Splatting
& \cellcolor{orange}25.85 & \cellcolor{red}30.67 & \cellcolor{red}29.43 & \cellcolor{orange}25.17 & \cellcolor{orange}28.37 & \cellcolor{orange}30.22 & \cellcolor{red}33.10 & \cellcolor{red}28.95 & \cellcolor{red}25.69 & \cellcolor{red}28.60
\\
\hline
$SA\text{-}GS_{fil}(ours)$
&\cellcolor{red}30.12 &\cellcolor{yellow}25.07 &\cellcolor{orange}28.80 &\cellcolor{red}27.83	&\cellcolor{red}29.29 &\cellcolor{red}32.16 &\cellcolor{yellow}28.23	&\cellcolor{yellow}24.42 &\cellcolor{orange}25.12 &\cellcolor{orange}27.89
\end{tabularx}
\end{subtable}

\vspace{0.2cm}

\begin{subtable}{\linewidth}
\centering
\fontsize{7pt}{8.5pt}\selectfont
\begin{tabularx}{\linewidth}{l|>{\centering\arraybackslash}X>{\centering\arraybackslash}X>{\centering\arraybackslash}X>{\centering\arraybackslash}X>{\centering\arraybackslash}X>{\centering\arraybackslash}X>{\centering\arraybackslash}X>{\centering\arraybackslash}X>{\centering\arraybackslash}X|>{\centering\arraybackslash}X}
& \multicolumn{9}{c}{\textbf{SSIM $\uparrow$}}\\
& \textit{bonsai} & \textit{bicycle} & \textit{counter} & \textit{garden} & \textit{kitchen} & \textit{room} & \textit{stump} & \textit{flowers} & \textit{treehill} & \textit{Avg.}\\ 
\hline
3DGS
&\cellcolor{orange}0.769 &0.593 &0.771 &0.616	&0.720 &0.850 &0.582	&0.559 &0.603 &0.674
\\
3DGS(MS)
& 0.718 & \cellcolor{orange}0.873 & \cellcolor{orange}0.867 & \cellcolor{yellow}0.686 & \cellcolor{orange}0.764 & \cellcolor{orange}0.859 & \cellcolor{orange}0.914 & \cellcolor{orange}0.737 & \cellcolor{orange}0.709 & \cellcolor{orange}0.792
\\
Mip-Splatting
& \cellcolor{yellow}0.733 & \cellcolor{red}0.898 & \cellcolor{red}0.877 & \cellcolor{orange}0.704 & \cellcolor{yellow}0.747 & \cellcolor{yellow}0.828 & \cellcolor{red}0.921 & \cellcolor{red}0.779 & \cellcolor{red}0.728 & \cellcolor{red}0.802
\\
\hline
$SA\text{-}GS_{fil}(ours)$
&\cellcolor{red}0.878 &\cellcolor{yellow}0.705 &\cellcolor{yellow}0.862 &\cellcolor{red}0.716	&\cellcolor{red}0.790 &\cellcolor{red}0.910 &\cellcolor{yellow}0.748	&\cellcolor{yellow}0.673 &\cellcolor{yellow}0.707 &\cellcolor{yellow}0.777
\end{tabularx}
\end{subtable}

\vspace{0.2cm}

\begin{subtable}{\linewidth}
\centering
\fontsize{7pt}{8.5pt}\selectfont
\begin{tabularx}{\linewidth}{l|>{\centering\arraybackslash}X>{\centering\arraybackslash}X>{\centering\arraybackslash}X>{\centering\arraybackslash}X>{\centering\arraybackslash}X>{\centering\arraybackslash}X>{\centering\arraybackslash}X>{\centering\arraybackslash}X>{\centering\arraybackslash}X|>{\centering\arraybackslash}X}
& \multicolumn{9}{c}{\textbf{LPIPS $\downarrow$}}\\
& \textit{bonsai} & \textit{bicycle} & \textit{counter} & \textit{garden} & \textit{kitchen} & \textit{room} & \textit{stump} & \textit{flowers} & \textit{treehill} & \textit{Avg.}\\
\hline
3DGS
&\cellcolor{yellow}0.250 &0.316 &0.238 &0.292	&0.266 &0.207 &0.326	&0.336 &0.345 &0.286
\\
3DGS(MS)
& \cellcolor{orange}0.228 & \cellcolor{red}0.154 & \cellcolor{red}0.148 & \cellcolor{red}0.251 & \cellcolor{red}0.183 & \cellcolor{red}0.140 & \cellcolor{red}0.131 & \cellcolor{red}0.234 & \cellcolor{red}0.261 & \cellcolor{red}0.192
\\
Mip-Splatting
& 0.275 & \cellcolor{orange}0.170 & \cellcolor{orange}0.188 & \cellcolor{yellow}0.290 & \cellcolor{yellow}0.248 & \cellcolor{yellow}0.198 & \cellcolor{orange}0.165 & \cellcolor{orange}0.249 & \cellcolor{orange}0.302 & \cellcolor{orange}0.232
\\
\hline
$SA\text{-}GS_{fil}(ours)$
&\cellcolor{red}0.196 &\cellcolor{yellow}0.305 &\cellcolor{yellow}0.210 &\cellcolor{orange}0.281	&\cellcolor{orange}0.237 &\cellcolor{orange}0.191 &\cellcolor{yellow}0.281	&\cellcolor{yellow}0.318 &\cellcolor{yellow}0.325 &\cellcolor{yellow}0.260
\end{tabularx}
\end{subtable}
\vspace{1em}
\caption{\textbf{Single-scale Training and Multi-scale testing on the Mip-NeRF 360 Dataset.} For each scene, we report the arithmetic mean of each metric averaged over the 4 scales($1\times$, $2\times$, $4\times$, $8\times$) used in the dataset. (MS) means multi-scale training.}
\label{table:per_scene_360_zoomin}
\end{table}

\begin{table}
\centering
\begin{subtable}{\linewidth}
\centering
\fontsize{7pt}{8.5pt}\selectfont
\begin{tabularx}{\linewidth}{l|>{\centering\arraybackslash}X>{\centering\arraybackslash}X>{\centering\arraybackslash}X>{\centering\arraybackslash}X>{\centering\arraybackslash}X>{\centering\arraybackslash}X>{\centering\arraybackslash}X>{\centering\arraybackslash}X|>{\centering\arraybackslash}X}
& \multicolumn{9}{c}{\textbf{PSNR $\uparrow$}}\\
& \textit{chair} & \textit{drums} & \textit{ficus} & \textit{hotdog} & \textit{lego} & \textit{materials} & \textit{mic} & \textit{ship} & \textit{Avg.}\\
\hline
3DGS
& 25.04 & 22.09 & 26.64 & 28.43 & 26.12 & 26.55 & 28.01 & 25.51 & 26.05
\\
3DGS(MS)
& 30.47 & 26.16 & 29.96 & 34.92 & 30.91 & 30.41 & 32.97 & 31.45 & 30.91
\\
Mip-Splatting
& \cellcolor{orange}32.57 & \cellcolor{orange}26.74 & \cellcolor{orange}32.92 & \cellcolor{yellow}34.44 & \cellcolor{orange}33.39 & \cellcolor{orange}31.53 & \cellcolor{orange}35.69 & \cellcolor{orange}32.36 & \cellcolor{orange}32.45
\\
\hline
$SA\text{-}GS_{fil}(ours)$
& \cellcolor{yellow}32.13 & \cellcolor{yellow}26.71 & \cellcolor{yellow}32.48 & \cellcolor{orange}35.66 & \cellcolor{yellow}33.05 & 30.90 & 33.78 & 30.39 & \cellcolor{yellow}31.89
\\
$SA\text{-}GS_{int}(ours)$
& 31.38 & 26.23 & 32.32 & 33.73 & 32.33 & \cellcolor{yellow}31.00 & \cellcolor{yellow}34.84 & \cellcolor{yellow}31.41 & 31.65
\\
$SA\text{-}GS_{sup}(ours)$
& \cellcolor{red}37.89 & \cellcolor{red}28.87 & \cellcolor{red}36.34 & \cellcolor{red}39.40 & \cellcolor{red}38.59 & \cellcolor{red}33.01 & \cellcolor{red}39.64 & \cellcolor{red}34.70 & \cellcolor{red}36.06
\end{tabularx}
\end{subtable}

\vspace{0.2cm}

\begin{subtable}{\linewidth}
\centering
\fontsize{7pt}{8.5pt}\selectfont
\begin{tabularx}{\linewidth}{l|>{\centering\arraybackslash}X>{\centering\arraybackslash}X>{\centering\arraybackslash}X>{\centering\arraybackslash}X>{\centering\arraybackslash}X>{\centering\arraybackslash}X>{\centering\arraybackslash}X>{\centering\arraybackslash}X|>{\centering\arraybackslash}X}
& \multicolumn{9}{c}{\textbf{SSIM $\uparrow$}}\\
& \textit{chair} & \textit{drums} & \textit{ficus} & \textit{hotdog} & \textit{lego} & \textit{materials} & \textit{mic} & \textit{ship} & \textit{Avg.}\\
\hline
3DGS
& 0.893 & 0.837 & 0.913 & 0.914 & 0.865 & 0.898 & 0.905 & 0.818 & 0.880
\\
3DGS(MS)
& 0.975 & 0.943 & 0.965 & \cellcolor{yellow}0.983 & 0.967 & 0.962 & 0.974 & \cellcolor{yellow}0.923 & 0.962
\\
Mip-Splatting
& \cellcolor{orange}0.984 & \cellcolor{yellow}0.950 & \cellcolor{orange}0.983 & \cellcolor{yellow}0.981 & \cellcolor{orange}0.979 & \cellcolor{orange}0.973 & \cellcolor{orange}0.981 & \cellcolor{orange}0.925 & \cellcolor{orange}0.970
\\
\hline
$SA\text{-}GS_{fil}(ours)$
& 0.977 & \cellcolor{orange}0.953 & \cellcolor{yellow}0.981 & \cellcolor{orange}0.985 & \cellcolor{yellow}0.975 & \cellcolor{yellow}0.972 & \cellcolor{orange}0.981 & 0.904 & \cellcolor{yellow}0.966
\\
$SA\text{-}GS_{int}(ours)$
& \cellcolor{yellow}0.978 & 0.943 & 0.981 & 0.978 & 0.974 & 0.970 & 0.977 & 0.919 & 0.965
\\
$SA\text{-}GS_{sup}(ours)$
& \cellcolor{red}0.994 & \cellcolor{red}0.968 & \cellcolor{red}0.991 & \cellcolor{red}0.992 & \cellcolor{red}0.993 & \cellcolor{red}0.980 & \cellcolor{red}\cellcolor{red}0.994 & \cellcolor{red}0.935 & \cellcolor{red}0.981
\end{tabularx}
\end{subtable}

\vspace{0.2cm}

\begin{subtable}{\linewidth}
\centering
\fontsize{7pt}{8.5pt}\selectfont
\begin{tabularx}{\linewidth}{l|>{\centering\arraybackslash}X>{\centering\arraybackslash}X>{\centering\arraybackslash}X>{\centering\arraybackslash}X>{\centering\arraybackslash}X>{\centering\arraybackslash}X>{\centering\arraybackslash}X>{\centering\arraybackslash}X|>{\centering\arraybackslash}X}
& \multicolumn{9}{c}{\textbf{LPIPS $\downarrow$}}\\
& \textit{chair} & \textit{drums} & \textit{ficus} & \textit{hotdog} & \textit{lego} & \textit{materials} & \textit{mic} & \textit{ship} & \textit{Avg.}\\
\hline
3DGS
& 0.047 & 0.090 & 0.059 & 0.040 & 0.073 & 0.051 & 0.044 & 0.124 & 0.066
\\
3DGS(MS)
& 0.022 & 0.057 & 0.033 & 0.019 & 0.034 & 0.036 & 0.027 & 0.083 & 0.039
\\
Mip-Splatting
& \cellcolor{orange}0.011 & \cellcolor{orange}0.042 & \cellcolor{orange}0.013 & \cellcolor{orange}0.014 & \cellcolor{orange}0.016 & \cellcolor{orange}0.020 & \cellcolor{orange}0.011 & \cellcolor{orange}0.072 & \cellcolor{orange}0.025
\\
\hline
$SA\text{-}GS_{fil}(ours)$
& 0.029 & 0.051 & 0.020 & 0.026 & 0.032 & 0.027 & 0.031 & 0.095 & 0.039
\\
$SA\text{-}GS_{int}(ours)$
& \cellcolor{yellow}0.016 & \cellcolor{yellow}0.046 & \cellcolor{yellow}0.014 & \cellcolor{yellow}0.018 & \cellcolor{yellow}0.021 & \cellcolor{yellow}0.021 & \cellcolor{yellow}0.015 & \cellcolor{yellow}0.077 & \cellcolor{yellow}0.029
\\
$SA\text{-}GS_{sup}(ours)$
& \cellcolor{red}0.006 & \cellcolor{red}0.035 & \cellcolor{red}0.008 & \cellcolor{red}0.010 & \cellcolor{red}0.007 & \cellcolor{red}0.017 & \cellcolor{red}0.006 & \cellcolor{red}0.066 & \cellcolor{red}0.019
\end{tabularx}
\end{subtable}
\vspace{1em}
\caption{\textbf{Single-scale Training and Multi-scale testing on the Blender Dataset.} For each scene, we report the arithmetic mean of each metric averaged over the 4 scales($1\times$, $\nicefrac{1}{2}\times$, $\nicefrac{1}{4}\times$, $\nicefrac{1}{8}\times$) used in the dataset. (MS) means multi-scale training.}
\label{table:per_scene_blender_zoomout}
\end{table}

\begin{table}
\centering
\begin{subtable}{\linewidth}
\centering
\fontsize{7pt}{8.5pt}\selectfont
\begin{tabularx}{\linewidth}{l|>{\centering\arraybackslash}X>{\centering\arraybackslash}X>{\centering\arraybackslash}X>{\centering\arraybackslash}X>{\centering\arraybackslash}X>{\centering\arraybackslash}X>{\centering\arraybackslash}X>{\centering\arraybackslash}X|>{\centering\arraybackslash}X}
& \multicolumn{9}{c}{\textbf{PSNR $\uparrow$}}\\
& \textit{chair} & \textit{drums} & \textit{ficus} & \textit{hotdog} & \textit{lego} & \textit{materials} & \textit{mic} & \textit{ship} & \textit{Avg.}\\
\hline
3DGS
& 24.26 & 22.14 & 23.72 & 27.87 & 23.99 & 25.45 & 29.22 & 26.86 & 25.44
\\
3DGS(MS)
& \cellcolor{red}30.47 & \cellcolor{red}26.16 & \cellcolor{red}29.96 & \cellcolor{red}34.92 & \cellcolor{red}30.91 & \cellcolor{red}30.41 & 32.97 & \cellcolor{red}31.45 & \cellcolor{red}30.91
\\
Mip-Splatting
& 30.11 & \cellcolor{orange}25.76 & \cellcolor{orange}28.51 & \cellcolor{orange}34.65 & \cellcolor{orange}29.54 & \cellcolor{orange}30.05 & \cellcolor{orange}33.79 & \cellcolor{orange}30.66 & \cellcolor{orange}30.39
\\
\hline
$SA\text{-}GS_{fil}(ours)$
& \cellcolor{orange}30.42 & \cellcolor{yellow}25.29 & \cellcolor{yellow}27.54 & \cellcolor{yellow}33.89 & \cellcolor{yellow}29.44 & \cellcolor{yellow}29.68 & \cellcolor{red}33.87 & \cellcolor{yellow}30.11 & \cellcolor{yellow}30.03
\end{tabularx}
\end{subtable}

\vspace{0.2cm}

\begin{subtable}{\linewidth}
\centering
\fontsize{7pt}{8.5pt}\selectfont
\begin{tabularx}{\linewidth}{l|>{\centering\arraybackslash}X>{\centering\arraybackslash}X>{\centering\arraybackslash}X>{\centering\arraybackslash}X>{\centering\arraybackslash}X>{\centering\arraybackslash}X>{\centering\arraybackslash}X>{\centering\arraybackslash}X|>{\centering\arraybackslash}X}
& \multicolumn{9}{c}{\textbf{SSIM $\uparrow$}}\\
& \textit{chair} & \textit{drums} & \textit{ficus} & \textit{hotdog} & \textit{lego} & \textit{materials} & \textit{mic} & \textit{ship} & \textit{Avg.}\\
\hline
3DGS
& 0.891 & 0.857 & 0.903 & 0.915 & 0.834 & 0.883 & 0.921 & 0.823 & 0.879
\\
3DGS(MS)
& \cellcolor{red}0.975 & \cellcolor{red}0.943 & \cellcolor{red}0.965 & \cellcolor{red}0.983 & \cellcolor{red}0.967 & \cellcolor{red}0.962 & \cellcolor{red}0.974 & \cellcolor{red}0.923 & \cellcolor{red}0.962
\\
Mip-Splatting
& \cellcolor{orange}0.951 & \cellcolor{orange}0.921 & \cellcolor{orange}0.947 & \cellcolor{orange}0.971 & \cellcolor{orange}0.930 & \cellcolor{orange}0.951 & \cellcolor{orange}0.965 & \cellcolor{orange}0.880 & \cellcolor{orange}0.939
\\
\hline
$SA\text{-}GS_{fil}(ours)$
& \cellcolor{yellow}0.949 & \cellcolor{yellow}0.912 & \cellcolor{yellow}0.933 & \cellcolor{yellow}0.967 & \cellcolor{yellow}0.925 & \cellcolor{yellow}0.946 & \cellcolor{orange}0.965 & \cellcolor{yellow}0.871 & \cellcolor{yellow}0.933
\end{tabularx}
\end{subtable}

\vspace{0.2cm}

\begin{subtable}{\linewidth}
\centering
\fontsize{7pt}{8.5pt}\selectfont
\begin{tabularx}{\linewidth}{l|>{\centering\arraybackslash}X>{\centering\arraybackslash}X>{\centering\arraybackslash}X>{\centering\arraybackslash}X>{\centering\arraybackslash}X>{\centering\arraybackslash}X>{\centering\arraybackslash}X>{\centering\arraybackslash}X|>{\centering\arraybackslash}X}
& \multicolumn{9}{c}{\textbf{LPIPS $\downarrow$}}\\
& \textit{chair} & \textit{drums} & \textit{ficus} & \textit{hotdog} & \textit{lego} & \textit{materials} & \textit{mic} & \textit{ship} & \textit{Avg.}\\
\hline
3DGS
& 0.071 & 0.106 & \cellcolor{yellow}0.068 & 0.073 & 0.122 & 0.084 & 0.063 & \cellcolor{yellow}0.146 & 0.092
\\
3DGS(MS)
& \cellcolor{red}0.022 & \cellcolor{red}0.057 & \cellcolor{red}0.033 & \cellcolor{red}0.019 & \cellcolor{red}0.034 & \cellcolor{red}0.036 & \cellcolor{red}0.027 & \cellcolor{red}0.083 & \cellcolor{red}0.039
\\
Mip-Splatting
& \cellcolor{orange}0.052 & \cellcolor{orange}0.091 & \cellcolor{orange}0.059 & \cellcolor{orange}0.048 & \cellcolor{orange}0.088 & \cellcolor{orange}0.060 & \cellcolor{orange}0.049 & \cellcolor{orange}0.138 & \cellcolor{orange}0.073
\\
\hline
$SA\text{-}GS_{fil}(ours)$
& \cellcolor{yellow}0.057 & \cellcolor{yellow}0.102 & 0.074 & \cellcolor{yellow}0.055 & \cellcolor{yellow}0.091 & \cellcolor{yellow}0.068 & \cellcolor{yellow}0.052 & 0.154 & \cellcolor{yellow}0.082
\end{tabularx}
\end{subtable}
\vspace{1em}
\caption{\textbf{Single-scale Training and Multi-scale testing on the Blender Dataset.} For each scene, we report the arithmetic mean of each metric averaged over the 4 scales($1\times$, $2\times$, $4\times$, $8\times$) used in the dataset. (MS) means multi-scale training.}
\label{table:per_scene_blender_zoomin}
\end{table}

\end{document}